\documentclass{article} %
    
\usepackage{etoolbox}
\newcommand{\arxiv}[1]{\iftoggle{alt}{}{#1}}
\newcommand{\alt}[1]{\iftoggle{alt}{#1}{}}
\newtoggle{alt}
\global\togglefalse{alt}

\alt{
  \usepackage{times} 
}

\arxiv{  
	\usepackage{natbib}
	\bibliographystyle{plainnat}
	\bibpunct{(}{)}{;}{a}{,}{,}
}

\arxiv{
	\usepackage[letterpaper, left=1in, right=1in, top=1in,
        bottom=1in]{geometry}
        \usepackage{hyperref}
        \hypersetup{colorlinks,citecolor=blue!70!black,linkcolor =
          blue!70!black}       %
        \usepackage{parskip}
      }

\arxiv{
  \usepackage[dvipsnames]{xcolor}         %
}

\usepackage[utf8]{inputenc} %
\usepackage[T1]{fontenc}    %
\usepackage{url}            %
\usepackage{booktabs}       %
\usepackage{amsfonts}       %
\usepackage{nicefrac}       %
\usepackage{microtype}      %
\arxiv{\usepackage{amsthm}}
\usepackage{mathtools}
\usepackage{amsmath}
\usepackage{bm}
\usepackage{bbm}
\usepackage{amssymb}
\usepackage{MnSymbol} %
\usepackage{makecell}
\usepackage{enumitem}
\usepackage{comment}
\usepackage{wrapfig}
\usepackage{colortbl}
\usepackage{setspace}
\usepackage{transparent}
\usepackage{inconsolata}
\usepackage[scaled=.90]{helvet}
\usepackage{xspace}
\usepackage{verbatim}
\usepackage{algorithm}
\usepackage[noend]{algpseudocode}

\newcommand{\multiline}[1]{\parbox[t]{\dimexpr\linewidth-\algorithmicindent}{#1}}
\usepackage{tablefootnote}
\usepackage{pifont}

\usepackage[nameinlink,capitalize]{cleveref}

\makeatletter
\newcommand{\neutralize}[1]{\expandafter\let\csname c@#1\endcsname\count@}
\makeatother

\usepackage{thmtools}
\arxiv{
\declaretheorem[name=Theorem,parent=section]{theorem}
\declaretheorem[name=Lemma,parent=section]{lemma}
\declaretheorem[name=Corollary,parent=section]{corollary}

\declaretheorem[name=Assumption, parent=section]{assumption}
\declaretheorem[name=Condition, parent=section]{condition}

\declaretheorem[name=Remark,parent=section]{remark}
\declaretheorem[name=Proposition, parent=section]{proposition}

}
\usepackage{crossreftools}
\makeatletter
\renewenvironment{proof}[1][Proof]%
{%
	\par\noindent{\bfseries\upshape {#1.}\ }%
}%
{\qed\newline}
\makeatother

\arxiv{
	\theoremstyle{plain}
\newtheorem{definition}[theorem]{Definition}
}

\usepackage{xpatch}
\xpatchcmd{\proof}{\itshape}{\normalfont\proofnameformat}{}{}
\newcommand{\proofnameformat}{\bfseries}

\renewcommand{\eqref}[1]{\texorpdfstring{\hyperref[#1]{(\ref*{#1})}}{(\ref*{#1})}}
\crefformat{equation}{#2Eq.\,(#1)#3}
\Crefformat{equation}{#2Eq.\,(#1)#3}

\Crefformat{figure}{#2Figure~#1#3}
\Crefformat{assumption}{#2Assumption~#1#3}

\Crefname{assumption}{Assumption}{Assumptions}

\def\ddefloop#1{\ifx\ddefloop#1\else\ddef{#1}\expandafter\ddefloop\fi}
\def\ddef#1{\expandafter\def\csname bb#1\endcsname{\ensuremath{\mathbb{#1}}}}
\ddefloop ABCDEFGHIJKLMNOPQRSTUVWXYZ\ddefloop
\def\ddefloop#1{\ifx\ddefloop#1\else\ddef{#1}\expandafter\ddefloop\fi}
\def\ddef#1{\expandafter\def\csname b#1\endcsname{\ensuremath{\mathbf{#1}}}}
\ddefloop ABCDEFGHIJKLMNOPQRSTUVWXYZ\ddefloop
\def\ddef#1{\expandafter\def\csname sf#1\endcsname{\ensuremath{\mathsf{#1}}}}
\ddefloop ABCDEFGHIJKLMNOPQRSTUVWXYZ\ddefloop
\def\ddef#1{\expandafter\def\csname c#1\endcsname{\ensuremath{\mathcal{#1}}}}
\ddefloop ABCDEFGHIJKLMNOPQRSTUVWXYZ\ddefloop
\def\ddef#1{\expandafter\def\csname h#1\endcsname{\ensuremath{\widehat{#1}}}}
\ddefloop ABCDEFGHIJKLMNOPQRSTUVWXYZ\ddefloop
\def\ddef#1{\expandafter\def\csname hc#1\endcsname{\ensuremath{\widehat{\mathcal{#1}}}}}
\ddefloop ABCDEFGHIJKLMNOPQRSTUVWXYZ\ddefloop
\def\ddef#1{\expandafter\def\csname t#1\endcsname{\ensuremath{\widetilde{#1}}}}
\ddefloop ABCDEFGHIJKLMNOPQRSTUVWXYZ\ddefloop
\def\ddef#1{\expandafter\def\csname tc#1\endcsname{\ensuremath{\widetilde{\mathcal{#1}}}}}
\ddefloop ABCDEFGHIJKLMNOPQRSTUVWXYZ\ddefloop
\def\ddefloop#1{\ifx\ddefloop#1\else\ddef{#1}\expandafter\ddefloop\fi}
\def\ddef#1{\expandafter\def\csname scr#1\endcsname{\ensuremath{\mathscr{#1}}}}
\ddefloop ABCDEFGHIJKLMNOPQRSTUVWXYZ\ddefloop

\newcommand{\veps}{\varepsilon}

\DeclareMathOperator*{\argmin}{arg\,min} %
\DeclareMathOperator*{\argmax}{arg\,max}

\def\ddef#1{\expandafter\def\csname b#1\endcsname{\ensuremath{\mb{#1}}}}
\ddefloop abcdeghijklmnopqrstuvwxyz\ddefloop

\newcommand{\mb}[1]{\boldsymbol{#1}}
\renewcommand{\bm}[1]{\boldsymbol{#1}}
\newcommand{\wt}[1]{\widetilde{#1}}

\let\underbar\undefined

\makeatletter
\let\save@mathaccent\mathaccent
\newcommand*\if@single[3]{%
	\setbox0\hbox{${\mathaccent"0362{#1}}^H$}%
	\setbox2\hbox{${\mathaccent"0362{\kern0pt#1}}^H$}%
	\ifdim\ht0=\ht2 #3\else #2\fi
}
\newcommand*\rel@kern[1]{\kern#1\dimexpr\macc@kerna}
\newcommand*\widebar[1]{\@ifnextchar^{{\wide@bar{#1}{0}}}{\wide@bar{#1}{1}}}
\newcommand*\underbar[1]{\@ifnextchar_{{\under@bar{#1}{0}}}{\under@bar{#1}{1}}}
\newcommand*\wide@bar[2]{\if@single{#1}{\wide@bar@{#1}{#2}{1}}{\wide@bar@{#1}{#2}{2}}}
\newcommand*\under@bar[2]{\if@single{#1}{\under@bar@{#1}{#2}{1}}{\under@bar@{#1}{#2}{2}}}
\newcommand*\wide@bar@[3]{%
	\begingroup
	\def\mathaccent##1##2{%
		\let\mathaccent\save@mathaccent
		\if#32 \let\macc@nucleus\first@char \fi
		\setbox\z@\hbox{$\macc@style{\macc@nucleus}_{}$}%
		\setbox\tw@\hbox{$\macc@style{\macc@nucleus}{}_{}$}%
		\dimen@\wd\tw@
		\advance\dimen@-\wd\z@
		\divide\dimen@ 3
		\@tempdima\wd\tw@
		\advance\@tempdima-\scriptspace
		\divide\@tempdima 10
		\advance\dimen@-\@tempdima
		\ifdim\dimen@>\z@ \dimen@0pt\fi
		\rel@kern{0.6}\kern-\dimen@
		\if#31
		\overline{\rel@kern{-0.6}\kern\dimen@\macc@nucleus\rel@kern{0.4}\kern\dimen@}%
		\advance\dimen@0.4\dimexpr\macc@kerna
		\let\final@kern#2%
		\ifdim\dimen@<\z@ \let\final@kern1\fi
		\if\final@kern1 \kern-\dimen@\fi
		\else
		\overline{\rel@kern{-0.6}\kern\dimen@#1}%
		\fi
	}%
	\macc@depth\@ne
	\let\math@bgroup\@empty \let\math@egroup\macc@set@skewchar
	\mathsurround\z@ \frozen@everymath{\mathgroup\macc@group\relax}%
	\macc@set@skewchar\relax
	\let\mathaccentV\macc@nested@a
	\if#31
	\macc@nested@a\relax111{#1}%
	\else
	\def\gobble@till@marker##1\endmarker{}%
	\futurelet\first@char\gobble@till@marker#1\endmarker
	\ifcat\noexpand\first@char A\else
	\def\first@char{}%
	\fi
	\macc@nested@a\relax111{\first@char}%
	\fi
	\endgroup
}
\newcommand*\under@bar@[3]{%
	\begingroup
	\def\mathaccent##1##2{%
		\let\mathaccent\save@mathaccent
		\if#32 \let\macc@nucleus\first@char \fi
		\setbox\z@\hbox{$\macc@style{\macc@nucleus}_{}$}%
		\setbox\tw@\hbox{$\macc@style{\macc@nucleus}{}_{}$}%
		\dimen@\wd\tw@
		\advance\dimen@-\wd\z@
		\divide\dimen@ 3
		\@tempdima\wd\tw@
		\advance\@tempdima-\scriptspace
		\divide\@tempdima 10
		\advance\dimen@-\@tempdima
		\ifdim\dimen@>\z@ \dimen@0pt\fi
		\rel@kern{0.6}\kern-\dimen@
		\if#31
		\underline{\rel@kern{-0.6}\kern\dimen@\macc@nucleus\rel@kern{0.4}\kern\dimen@}%
		\advance\dimen@0.4\dimexpr\macc@kerna
		\let\final@kern#2%
		\ifdim\dimen@<\z@ \let\final@kern1\fi
		\if\final@kern1 \kern-\dimen@\fi
		\else
		\underline{\rel@kern{-0.6}\kern\dimen@#1}%
		\fi
	}%
	\macc@depth\@ne
	\let\math@bgroup\@empty \let\math@egroup\macc@set@skewchar
	\mathsurround\z@ \frozen@everymath{\mathgroup\macc@group\relax}%
	\macc@set@skewchar\relax
	\let\mathaccentV\macc@nested@a
	\if#31
	\macc@nested@a\relax111{#1}%
	\else
	\def\gobble@till@marker##1\endmarker{}%
	\futurelet\first@char\gobble@till@marker#1\endmarker
	\ifcat\noexpand\first@char A\else
	\def\first@char{}%
	\fi
	\macc@nested@a\relax111{\first@char}%
	\fi
	\endgroup
}
\makeatother

\newcommand{\alghyperref}[1]{\hyperref[#1]{Alg.~\ref*{#1}}}

\newcommand{\inner}[2]{\langle #1,#2\rangle}
\newcommand{\tr}{\mathrm{Tr}}

\newcommand{\what}{\widehat}

\renewcommand{\ln}{\log}        %

\let\oldparagraph\paragraph
\renewcommand{\paragraph}[1]{\oldparagraph{#1.}}
\renewcommand{\colon}{:}        %

\newcommand{\reals}{\mathbb{R}}

\newcommand{\E}{\mathbb{E}}

\newcommand{\nn}{\nonumber}

\newcommand{\bxi}{\bm{\xi}}
\newcommand{\x}{\bm{x}}
\newcommand{\z}{\bm{z}}

\newcommand{\y}{\bm{y}}

\newcommand{\wtilde}[1]{\widetilde{#1}}

\newcommand{\algcommentlight}[1]{\textcolor{blue!70!black}{\transparent{0.5}\footnotesize{\texttt{\textbf{//\hspace{2pt}#1}}}}}

\newcommand{\algcommentbiglight}[1]{\textcolor{blue!70!black}{\transparent{0.5}\footnotesize{\texttt{\textbf{/* #1~*/}}}}}

\newcommand{\bigoht}{\wt{O}}

\newcommand{\polylog}{\mathrm{polylog}}

\newcommand{\norm}[1]{\|#1 \|}

\newcommand{\g}{\bm{g}}
\newcommand{\w}{\bm{w}}
\newcommand{\K}{\mathcal{K}}
\renewcommand{\u}{\bm{u}}

\newcommand{\sep}{\mathrm{sep}}
\newcommand{\Bfrak}{\mathfrak{B}}
\newcommand{\Gtilde}{\wtilde{G}}
\newcommand{\gt}{\wtilde\g}
\newcommand{\vv}{\bm{v}}

\newcommand{\ons}{\texttt{ONS}}

\newcommand{\bnabla}{\bm{\nabla}}
\newcommand{\utilde}{\wtilde{\u}}
\newcommand{\inte}{\mathrm{int}}
\newcommand{\ftrl}{\texttt{FTRL}}

\newcommand{\bnablat}{\wtilde\bnabla}

\newcommand{\dom}{\mathrm{dom}\,}

\newcommand{\barons}{\texttt{Barrier}\text{-}\texttt{ONS}}

\newcommand{\Sep}{\mathrm{Sep}}
\newcommand{\gb}{\bar{\g}}
\newcommand{\Reg}{\texttt{Reg}}
\newcommand{\gproj}{\g^\star}
\newcommand{\wproj}{\w^\star}
\newcommand{\gauged}{\texttt{GaugeDist}}
\newcommand{\bdelta}{\bm{\delta}} 

\Crefformat{line}{#2Line #1#3}

\makeatletter
\let\OldStatex\Statex
\renewcommand{\Statex}[1][3]{%
  \setlength\@tempdima{\algorithmicindent}%
  \OldStatex\hskip\dimexpr#1\@tempdima\relax}
\makeatother

\newcommand{\fakepar}[1]{\paragraph{#1}}

\usepackage{color-edits}

\addauthor{zm}{red}
\newcommand{\zm}[1]{\zmcomment{#1}}
\usepackage{autonum}

\arxiv{
  \title{Online Convex Optimization with a Separation Oracle}
  \author{Zakaria Mhammedi\\{\small \texttt{mhammedi@google.com}}}
\date{}
}

\alt{
\title[Online Convex Optimization with a Separation Oracle]{Online Convex Optimization with a Separation Oracle}
}

\begin{document}
	\maketitle
	\begin{abstract}
In this paper, we introduce a new projection-free algorithm for Online Convex Optimization (OCO) with a state-of-the-art regret guarantee among separation-based algorithms. Existing projection-free methods based on the classical Frank-Wolfe algorithm achieve a suboptimal regret bound of $O(T^{3/4})$, while more recent separation-based approaches guarantee a regret bound of $O(\kappa \sqrt{T})$, where $\kappa$ denotes the asphericity of the feasible set, defined as the ratio of the radii of the containing and contained balls. However, for ill-conditioned sets, $\kappa$ can be arbitrarily large, potentially leading to poor performance. Our algorithm achieves a regret bound of $\wtilde{O}(\sqrt{dT} + \kappa d)$, while requiring only $\wtilde{O}(1)$ calls to a separation oracle per round. Crucially, the main term in the bound, $\wtilde{O}(\sqrt{d T})$, is independent of $\kappa$, addressing the limitations of previous methods. Additionally, as a by-product of our analysis, we recover the $O(\kappa \sqrt{T})$ regret bound of existing OCO algorithms with a more straightforward analysis and improve the regret bound for projection-free online exp-concave optimization. Finally, for constrained stochastic convex optimization, we achieve a state-of-the-art convergence rate of $\wtilde{O}(\sigma/\sqrt{T} + \kappa d/T)$, where $\sigma$ represents the noise in the stochastic gradients, while requiring only $\wtilde{O}(1)$ calls to a separation oracle per iteration. 	\end{abstract}

  \alt{
  }

	\tableofcontents

	\section{Introduction}
	\label{sec:intro}
Convex optimization is a foundational tool in computer science and machine learning, underpinning many modern techniques in these fields. Although classical algorithms like interior-point and cutting-plane methods are effective \citep{grotschel2012,bubeck2015,lee2018efficient,lee2019solving}, they become computationally prohibitive as problem sizes and dimensions grow. Given the high-dimensional nature of many contemporary problems, there is a growing demand for alternative algorithms that preserve strong theoretical guarantees while being computationally efficient enough to tackle large-scale optimization tasks.

Online Gradient Descent (OGD) \citep{zinkevich2003online} is a popular first-order optimization method that trades a higher number of iterations for lower memory usage and per-step computational cost, making it widely used in practice. However, in constrained convex optimization, OGD requires a Euclidean projection onto the feasible set at every step in the worst-case, which can be computationally expensive, especially for complex feasible sets. This drawback often offsets the benefits of first-order methods. To address this, projection-free optimization methods have been developed \citep{hazan2008sparse,jaggi2013,lacoste2015global,garber2016}, with the most well-known being the Frank-Wolfe algorithm \citep{frank1956}, which replaces costly Euclidean projections with more efficient \emph{linear optimization} over the feasible set. More recently, another class of projection-free algorithms has emerged that uses \emph{membership} or \emph{separation} oracles instead of linear optimization \citep{mhammedi2022efficient,garber2022new,lu2023projection,grimmer2024radial}, providing greater flexibility and efficiency in handling constraints.

Most modern projection-free algorithms are designed for Online Convex Optimization (OCO) \citep{hazan2016introduction}, a framework that generalizes classical offline convex optimization. In OCO, at each round $t$, the algorithm selects a vector \(\w_t\) from the feasible set $\K \subset \reals^d$ and incurs a loss \(f_t(\w_t)\), where $f_t$ is a convex function that may be adversarially chosen. The objective is to ensure that the regret $\Reg_T \coloneqq \sup_{\w \in \K} \sum_{t=1}^T \left( f_t(\w_t) - f_t(\w) \right)$ grows sublinearly in $T$. Existing projection-free algorithms, such as those in \citep{hazan2012, mhammedi2022efficient, garber2022new}, achieve sublinear regret while requiring only \(\wtilde{O}(1)\) calls per round to a linear optimization or separation oracle. Sublinear regret in OCO translates into guarantees for offline and stochastic convex optimization via standard online-to-batch conversion techniques \citep{cesa2006prediction, shalev2011, cutkosky2019}, where smaller regret yields better convergence rates. While (projected) OGD achieves an optimal, dimension-free regret of $O(\sqrt{T})$,\footnote{See, e.g., \citep{abernethy2008optimal} for a lower bound.} there remains a significant gap between this bound and the regret bounds of state-of-the-art projection-free algorithms \citep{hazan2012, mhammedi2022efficient, garber2022new, lu2023projection}. In this paper, we aim to close that gap.
 
The current state-of-the-art regret bound for linear optimization-based algorithms (e.g., Frank-Wolfe-style algorithms) is $O(T^{3/4})$ \citep{hazan2012}. Since this result was first established, no improvements have been made without introducing additional structural assumptions on the objective function or feasible set. It remains an open question whether this is the best achievable regret bound for online algorithms that make only a constant number of calls to a linear optimization oracle per round. More recently, \cite{mhammedi2022efficient, garber2022new} introduced a new class of projection-free algorithms that use separation oracles instead of linear optimization oracles, guaranteeing a regret bound of $O(\kappa \sqrt{T})$, where $\kappa \coloneqq \frac{R}{r}$ represents the \emph{asphericity} of the set $\K$, defined as the ratio between the radii of the containing and contained balls. Achieving this $O(\kappa \sqrt{T})$ regret bound, which provides optimal dependence on the number of rounds $T$, was somewhat surprising given the difficulty of improving the $O(T^{3/4})$ bound for linear optimization-based algorithms. However, the asphericity factor $\kappa$ can be arbitrarily large for ill-conditioned feasible sets, which poses a challenge. While previous work has shown that for many sets of interest, $\kappa$ is $O(d^\alpha)$ with $\alpha < 1$ \citep{mhammedi2022efficient}, and any convex set can be pre-processed (e.g., put in isotropic position) to ensure $\kappa \leq d$ \citep{flaxman2005online}, this results in a potentially high pre-processing computational cost and a worst-case regret bound of $O(d\sqrt{T})$. In this paper, we show that the dependence on $\kappa$ and the dimension $d$ in the regret bounds for separation-based algorithms can be further improved. 

\paragraph{Contributions} In this paper, we introduce a separation-based projection-free algorithm for OCO that improves upon the state-of-the-art guarantees for such algorithms. Specifically, our method achieves a regret bound of $\wtilde{O}(\sqrt{dT} + \kappa d)$ while making only $\wtilde{O}(1)$ calls to a separation oracle per round. Crucially, the main term of this bound, $\wtilde{O}(\sqrt{d T})$, is \emph{independent} of the asphericity $\kappa$. As discussed earlier, existing separation-based algorithms can incur regret as large as $\wtilde{O}(d \sqrt{T})$ in the worst case, even after preprocessing the feasible set. Our bound improves this by a factor of $\sqrt{d}$, without any need for preprocessing.

 As a by-product of our analysis, we provide an improved regret bound for projection-free online exp-concave optimization, which we required in an intermediate step of our analysis. Additionally, we recover the $O(\kappa \sqrt{T})$ regret bound for OCO with a simplified analysis.

 By applying a standard online-to-batch conversion, our new $\wtilde{O}(\sqrt{dT})$ regret translates to a $\wtilde{O}(\sqrt{{d}/{T}})$ convergence rate in offline and stochastic optimization settings. Additionally, by leveraging our intermediate results on projection-free exp-concave optimization (which are of independent interest), we achieve a faster convergence rate of $\wtilde{O}(\sigma \sqrt{d/T} + {\kappa d}/T)$. Notably, this rate simplifies to $\wtilde{O}(\kappa d/T)$ when $\sigma = 0$ (i.e., in the offline setting). Our results are summarized in \cref{tb:resultscomp}.

\begin{table}[tp]
    \caption{Comparison of projection-free regret bounds for OCO (see \cref{sec:setup2}) and Stochastic Convex Optimization (SCO) (see \cref{sec:stochastic}). Here, $\kappa \coloneqq R/r$ represents the asphericity of the feasible set $\cK$, where $r$ and $R$ are such that $\bbB(r) \subseteq \K \subseteq \bbB(R)$. For ill-conditioned sets, $\kappa$ can be arbitrarily large. Unlike existing regret bounds, our new bound places $\kappa$ in a lower-order term, rather than having it multiply $\sqrt{T}$. In SCO, $\sigma^2$ represents the variance of the stochastic gradients. In the offline setting (i.e., $\sigma=0$), our method achieves the fast rate of $O(\frac{\kappa d}{T})$.
}
\label{tb:resultscomp}
\renewcommand{\arraystretch}{1.6}
\fontsize{9}{10}\selectfont
\centering 
\begin{tabular}{ccccc}
\hline
 Papers & \makecell{Regret bound\\ in OCO} &  \makecell{Convergence rate\\ in SCO}  & Oracle type &                                 
\makecell{Number of oracle \\ calls per round}  \\
\hline
 \citep{hazan2012}& $O(T^{3/4})$ & $O\left(\frac{1}{T^{1/3}}\right)$ &  Linear optimization & 1  \\
 \hline
 \makecell{\citep{mhammedi2022efficient} \\ \citep{garber2022new}} & $O(\bm{\kappa} \sqrt{T})$ & $O\left(\frac{\bm{\kappa}}{\sqrt{T}}\right)$ & Separation  & $O(1)$-$\wtilde{O}(1)$  \\
 \hline
{\bf This paper}  &
$\wtilde{O}(\sqrt{dT} + \bm{\kappa} d)$ & $\wtilde{O}\left(\bm\sigma \sqrt{\frac{d}{T}} + \frac{\bm{\kappa} d}{T}\right)$ & Separation & $\wtilde{O}(1)$\\
\hline
\end{tabular}
\end{table}

\paragraph{Related works}
Our approach builds on the projection-free reduction method introduced by \cite{mhammedi2022efficient}, which transforms any OCO problem over a feasible set $\K$ into an OCO problem over a ball $\bbB(R)$ containing $\K$ (i.e., $\K \subseteq \bbB(R)$), where Euclidean projections can be computed at a cost of $O(d)$. This method is closely related to the earlier ‘constrained-to-unconstrained’ reduction by \cite{cutkosky2018blackbox}, but instead of relying on potentially expensive Euclidean projections onto $\K$, it uses Gauge projections, which can be performed efficiently with a logarithmic number of calls to a separation oracle for the feasible set. The reduction in \cite{mhammedi2022efficient} has also been successfully applied outside the OCO setting, providing global non-asymptotic superlinear rates for a quasi-Newton method \citep{jiang2023online,jiang2024accelerated}, and for the design of extension functions in bandit convex optimization \citep{fokkema2024online}.

Our approach integrates the projection-free reduction from \citep{mhammedi2022efficient} with the efficient exp-concave optimization algorithm introduced by \cite{mhammedi2023quasi}. The latter is an Online Newton Step (\ons) method that implicitly tracks specific Follow-The-Regularized-Leader (\ftrl) iterates, where the associated regularizer is the log-barrier for a Euclidean ball. By exploiting the unique structure of the log-barrier for the Euclidean ball, \cite{mhammedi2023quasi} show that the \emph{generalized projections},\footnote{Also known as Mahalanobis projection; see e.g.~\cite{mhammedi2019lipschitz}.} typically required by the classical \ons{} algorithm \citep{hazan2007logarithmic} and often computationally expensive even for a ball \citep{koren2013open}, can be completely bypassed. Additionally, they prove that the algorithm requires only $\wtilde{O}(1)$ calls to a separation oracle per round, along with $\wtilde{O}(1)$ matrix-vector multiplications. While our primary focus is on OCO, we incorporate several of the techniques in \cite{mhammedi2023quasi} and, somewhat unexpectedly, achieve state-of-the-art guarantees in OCO by taking a detour through online exp-concave optimization.

\paragraph{Outline} In \cref{sec:setup}, we describe the OCO setup and introduce the necessary notation and definitions. In \cref{sec:ons}, we present \barons, an efficient projection-free algorithm for exp-concave optimization over a ball, which forms a key component of our final method for OCO. In \cref{sec:oco}, we present our results for OCO and show how we reduce the problem to online exp-concave optimization over a ball. In \cref{sec:stochastic}, we extend these results to the stochastic and offline convex optimization settings, achieving a state-of-the-art convergence rate for projection-free stochastic convex optimization. We conclude with a discussion of our results and future work in \cref{sec:conclusion}. 	
	\section{Preliminaries}
	\label{sec:setup}
In \cref{sec:setup}, we formally introduce the OCO setup and the notation used. In \cref{sec:gaugedist}, we present key convex analysis notations and preliminary results that are used throughout the paper.
  
\subsection{Setup and Notation}
\label{sec:setup2}
Throughout, let $\K$ denote a closed convex subset of the Euclidean space $\reals^d$. We consider the standard OCO setup over $\K$, where an algorithm produces iterates within $\K$ over multiple rounds. At the start of each round $t \geq 1$, the algorithm outputs $\w_t$ and incurs a loss $f_t(\w_t)$, where $f_t: \K \rightarrow \reals$ is a convex function, potentially chosen adversarially based on the history and $\w_t$. As is typical in OCO literature, we assume that the algorithm observes only a subgradient $\g_t \in \partial f_t(\w_t)$, rather than the full function. The algorithm’s performance is evaluated in terms of regret after $T \geq 1$ rounds:
\begin{align}
 \Reg_T \coloneqq \sum_{t=1}^T f_t(\w_t)- \inf_{\w\in \K} \sum_{t=1}^T f_t(\w).
\end{align} 
By the convexity of $(f_t)$, $\Reg_T$ is bounded from above by the \emph{linearized regret}: $\sum_{t=1}^T \inner{\g_t}{\w_t} - \inf_{\w \in \K} \sum_{t=1}^T \inner{\g_t}{\w}$. Therefore, to bound $\Reg_T$, it is sufficient to bound the linearized regret.

Our goal in this paper is to design an efficient OCO algorithm that achieves sublinear regret while requiring only a logarithmic number of calls per round to a separation oracle for the feasible set $\K$, rather than relying on Euclidean projections.
\begin{definition}[Separation oracle]
  \label{def:separation}
  A \emph{Separation oracle} $\Sep_{\cC}$ for a set $\cC$ is an oracle that given $\w \in \reals^d$ returns a pair $(b,\vv) \in \{0,1\} \times \bbB(1)$ (where $\bbB(1)$ denotes the unit Euclidean ball in $\reals^d$), such that 
  \begin{itemize}
    \item $b =0$ and $\vv=\bm{0}$, if $\w \in \cC$; and otherwise,
    \item $b =1$ and $\inner{\vv}{\w}> \inner{\vv}{\u}$, for all $\u\in \K$.
  \end{itemize}
  We denote by $C_{\sep}(\cC)$ the computational cost of one call to this oracle. 
\end{definition}
We consider the OCO problem described above and additionally assume that the functions $(f_t)$ are $G$-Lipschitz, for some $G>0$, and that $\K$ is ``sandwiched'' between two Euclidean balls with radii $r$ and $R$. To formalize these assumptions, let $\|\cdot\|$ denote the Euclidean norm, and $\bbB(\gamma) \subset \reals^d$ represent the Euclidean ball of radius $\gamma > 0$.

\begin{assumption}
  \label{ass:setassumption}
  The set $\K \subseteq \reals^d$ is a closed and convex and there are some $r,R>0$ such that \[\bbB(r) \subseteq \K \subseteq \bbB(R).\] 
\end{assumption}

\begin{assumption}
  \label{ass:online}
  There is some $G>0$, such that for all $t\geq 1$, the function $f_t: \K \rightarrow \reals$ is convex and for all $\w\in \K$ and $\g_t \in \partial f_t(\w)$, we have $\|\g_t\|\leq G$.
\end{assumption}

\paragraph{Additional notation}
We denote by $\K^\circ\coloneqq \{\x \in \reals^d\colon  \inner{\x}{\y} \leq 1, \forall \y \in \K \}$ the \emph{polar} set of $\K$ \citep{hiriart2004}. We denote by $\inte\, \K$ the interior of a set $\K$. Given a function $f:\cC \rightarrow \reals$ on a compact set $\cC$, we let $\argmin_{\x\in \cC} f(\x)$ denote the subset of points in $\cC$ that minimize the function $f$.
We use $\bigoht(\cdot)$ to denote a bound up to factors polylogarithmic in parameters appearing in the expression.

\subsection{Gauge Distance and Projection}
\label{sec:gaugedist}
We now introduce some convex analysis concepts and preliminary results that will be used throughout the paper, beginning with the notion of a Gauge function (a.k.a.~Minkowski functional \citep{hiriart2004}). In this section, let $\cC \subseteq \reals^d$ represent a closed convex set.
\begin{definition}
\label{def:gaugefunction}
The Gauge function $\gamma_\cC:\reals^d \rightarrow \reals$ of the set $\cC$ is defined as 
\begin{align}
  \gamma_\cC(\u) \coloneqq \inf\{\lambda \in \reals_{\geq 0} \mid \u \in \lambda \cC \}.
\end{align}
\end{definition}
The Gauge function $\gamma_{\cC}$ can be viewed as a ``pseudo'' norm induced by the convex set $\cC$; it becomes a true norm when $\cC$ is centrally symmetric (i.e., $\cC = -\cC$). With the Gauge function defined, we can introduce the Gauge distance \citep{mhammedi2022efficient}, a key concept in the approach of this paper.
\begin{definition}[Gauge distance]
\label{def:gaugedist}
The Gauge distance function $S_\cC$ corresponding to the set $\cC$ is defined as 
\begin{align}
  \forall \u \in \reals^d,\quad S_{\cC}(\u) \coloneqq \inf_{\x \in \cC} \gamma_{\cC}(\u - \x).
\end{align}
\end{definition}
This naturally leads to the concept of the Gauge projection \citep{mhammedi2022efficient}.
\begin{definition}
  \label{def:gaugeproj}
  The Gauge projection operator $\Pi^{\mathrm{gau}}_\cC$ induced by $\cC$ is the set-valued mapping defined as:
  \begin{align}
    \forall \u \in \reals^d, \quad \Pi^{\mathrm{gau}}_\cC (\u)\coloneqq \argmin_{\x\in \cC} \gamma_\cC(\u - \x).
  \end{align}
\end{definition}
\begin{algorithm}[t]
	\caption{$\gauged{}(\w;\cC,\veps,r)$: Approximate value and subgradient of the Gauge distance function.}  
	\label{alg:gauge}
	\begin{algorithmic}[1]
		\Statex[0] {\bf require:} Separation oracle $\Sep_{\cC}$ for $\cC$, input vector $\w \in \reals^d$, and parameters $\veps,r>0$.
        \Statex[0] {\bf returns} $S \approx S_{\cC}(\w)$ and $\bs \approx \partial S_{\cC}(\w)$, where $S_{\cC}(\u) \coloneqq \inf_{\x\in \cC} \gamma_{\cC}(\u -\x)$ is the Gauge distance function.
        \State Set $(b, \vv) \gets \Sep_\cC(\w)$. \hfill \algcommentlight{$b=1$ if $\w \in \cC$ and $0$, otherwise.}
		\If{$b=1$}  \hfill{\algcommentlight{This corresponds to the case where $\w\in \cC$.}}  \label{line:memset}
        \State Set $(S,\bs)\gets (0,\bm{0})$.
		\State {\bf return} $(S,\bs)$.\label{line:memsetnext}
		\EndIf
		\State Set $\alpha \gets 0$, $\beta\gets 1$, and $\mu\gets(\alpha+\beta)/2$. 
		\While{$\beta-\alpha> \frac{r^2 \veps}{2\|\w\|^2}$} \label{line:while}
        \State Set $(b, \vv)\gets \Sep_\cC(\mu \w)$. \hfill \algcommentlight{$b=1$ if $\mu\w \in \cC$ and $0$, otherwise.}
		\State Set $\alpha\gets\mu$ if $b=1$; and $\beta\gets\mu$ otherwise.\label{line:cond}
		\State Set $\mu\gets(\alpha+\beta)/2$.
		\EndWhile
        \State Set $S\gets \alpha^{-1} -1$ and $\bs \gets \frac{\vv}{\beta \cdot \vv^\top \w}$.
		\State {\bf return} $(S,\bs)$. 
	\end{algorithmic}
\end{algorithm} Our projection-free OCO approach in this paper leverages Gauge projections instead of standard Euclidean projections. As we will soon demonstrate, Gauge projections can be performed efficiently using a separation oracle. To understand this, we first need the following result from \citep{mhammedi2022efficient}, which allows us to express both the Gauge distance and its subgradients in terms of the Gauge function $\gamma_\cC$.
\begin{lemma}
  \label{lem:gauge}
  Suppose that the set $\cC$ satisfies $\bm{0}\in \inte\,\cC$. Then, for any $\w\in \reals^d$, we have 
  \begin{gather} 
           \Pi^{\mathrm{gau}}_{\cC}(\w)= \left\{\begin{array}{ll} \frac{\w}{\gamma_{\cC}(\w)}, & \text{if } \w\not\in \cC; \\ \w,& \text{otherwise}.    \end{array}   \right.
          \shortintertext{and the Gauge distance function satisfies:}
S_\cC(\w) = \max(0,\gamma_{\cC}(\w) -1); \ \ \text{and} \ \  \partial S_\cC(\w) = \left\{ \begin{array}{ll} \partial \gamma_{\cC}(\w)=  \argmax_{\x\in \cC^\circ}\inner{\x}{\w}, & \text{if } \w\notin \cC;\\ \{ \bm{0}\},& \text{otherwise}.  \end{array}   \right. \label{eq:dist}
  \end{gather} 
\end{lemma}
The key implication of \cref{lem:gauge} is that, to compute the Gauge distance and its subgradients---both of which are needed for our OCO algorithm---it is sufficient to compute or approximate the Gauge function $\gamma_\cC$ and its subgradients. This is accomplished through \cref{alg:gauge}, whose guarantee we now present (with the proof provided in \cref{sec:subgradient}). 
\begin{lemma}
	\label{lem:approxgrad}
	Let $\veps,r>0$ and $\w\in \reals^d$ be given, and suppose that $\bbB(r)\subseteq \cC$. Consider a call to \cref{alg:gauge} with input $(\cC,\w, \veps, r)$. Then, the output $(S,\bs)$ of \cref{alg:gauge} satisfies 
	\begin{align}
    \|\bs\|\leq 1/r, \qquad  S_\cC(\w) \leq S \leq S_\cC(\w)  + \veps, \quad \text{and}\quad  
	 \forall \u \in \reals^d,\quad S_\cC(\u) \geq S_\cC(\w) + (\u - \w)^\top \bs -\veps. \label{eq:pand}
	\end{align}
	Furthermore, \cref{alg:gauge} makes at most $1+ \log_2(\frac{4 \|\w\|^2}{r^2\veps})$ calls to the separation oracle $\Sep_{\cC}$ in \cref{def:separation}.
	\end{lemma}

       \label{sec:main}

\section{Barrier-Regularized Online Newton Steps over a Ball}
\label{sec:ons}
In this section, we present \barons{} (\cref{alg:pseudo}), a key component of our projection-free OCO approach detailed in the next section. \barons{} generates online Newton iterates that implicitly track specific Follow-The-Regularized-Leader (\ftrl) iterates, where the associated regularizer is the log-barrier for a Euclidean ball.

\begin{algorithm}[h]
    \caption{\barons: Barrier-regularized \ons{} over a Euclidean ball. Pseudocode of \cref{alg:fastexpconcave}.}
	\label{alg:pseudo} 
	\begin{algorithmic}[1]
       \arxiv{ \Statex[0]  {\bf inputs:} Number of rounds $T\geq1$, and parameters $\eta,\nu, c >0$.}
       \alt{\Statex  {\bf inputs:} Parameters $\eta,\nu, c >0$}
        \State Set $\z_1\gets \bm{0}$, $\u_1\gets \bm{0}$, and $m\gets \mathfrak{c}\cdot \log_c(dT)$, with $\mathfrak{c}$ a sufficiently large universal constant.
                \For{$t=1,\dots, T$}
        \State Play $\u_t$ and observe $\gt_t$.%
    \State Set $\Sigma_{t} \gets \left(\frac{2 \nu  I}{R^2-\|\z_t\|^2}+ \frac{4\nu \u_t \u_t^\top}{(R^2-\|\u_t\|^2)^2}   + \eta \sum_{s=1}^t \gt_s \gt_s^\top \right)^{-1}$.\hfill \algcommentlight{Can be computed in $O(d^2)$ most rounds---see \cref{alg:fastexpconcave}} \label{line:expense} %
        \State Set $\gamma_t\gets \frac{2\nu}{R^2-\|\z_t\|^2}- \frac{2\nu}{R^2-\|\u_t\|^2}$.%
        \State Set $H_t \gets \sum_{k=1}^{m+1}\gamma_t^{k-1}  \Sigma_t^{k}$. \hfill   \algcommentlight{Taylor approximation of $\nabla^{-2}\Phi_{t+1}(\u_t)$}
        \State Set $\u_{t+1}\gets\u_t -  H_t \nabla\Phi_{t+1}(\u_t)$. \hfill \algcommentlight{Approximate Newton step.} \label{line:taylor} %
       \arxiv{ \Statex[1] \algcommentbiglight{Check if the Taylor approximation point needs to be updated}}
       \alt{ \Statex \algcommentbiglight{Check if the Taylor expansion point needs to be updated}}
       \If{$|\|\u_{t+1}\|^2-\|\z_{t}\|^2| \leq  c \cdot (R^2-\|\z_t\|^2)$}  \label{line:mark}
        \State Set $\z_{t+1}\gets\z_{t}$. %
        \Else 
        \State Set $\z_{t+1}\gets\u_{t+1}$.\label{line:endmark}%
   \EndIf  
   \EndFor
	\end{algorithmic}
	\end{algorithm} \barons{} was originally proposed by \cite{mhammedi2023quasi} in the context of online and stochastic exp-concave optimization as a more computationally efficient alternative to the classical Online Newton Step (\texttt{ONS}) algorithm \citep{hazan2007logarithmic}. Unlike \texttt{ONS}, \barons{} eliminates the need for \emph{generalized projections} onto the feasible set, which can be computationally expensive, even for a Euclidean ball. This is achieved by generating iterates that remain close (in Euclidean distance) to the \ftrl{} iterates, which stay within the interior of the feasible set due to the log-barrier. In this paper, we adopt \barons{} from \citep{mhammedi2023quasi} with only minor modifications, while improving its analysis and guarantees for the online exp-concave optimization setting. 

We now provide a brief description of \barons{} (\cref{alg:pseudo}); for further details, the reader may refer to \citep{mhammedi2023quasi}. \cref{alg:pseudo} offers a simplified version of \cref{alg:fastexpconcave}, written for ease of understanding. While \cref{alg:pseudo} is technically equivalent to \cref{alg:fastexpconcave}, it abstracts away the details of how certain steps can be implemented efficiently. \cref{alg:fastexpconcave}, on the other hand, makes these efficiency considerations explicit. In the following discussion, we will focus on \cref{alg:pseudo}.

\subsection{\barons: Algorithm Description} To describe \barons{} (\cref{alg:pseudo}), we first need to introduce a series of objective functions $(\Phi_t)$. Given parameters $\eta, \nu, R > 0$ and the history of observed loss vectors $(\gt_s)_{s<t}$ before round $t\geq 1$, the objective function $\Phi_t$ is defined as:
\begin{align}
	\Phi_t(\x) \coloneqq \Psi(\x)  + \frac{\eta}{2} \sum_{s=1}^{t-1} \inner{\gt_s}{\x - \u_s}^2 + \x^\top \sum_{s=1}^{t-1} \gt_s, \quad \text{where} \quad \Psi(\x) \coloneqq -\nu\log(R^2 - \|\x\|^2). \label{eq:Phi}
\end{align} 
Note that $\Psi/\nu$ is the standard log-barrier for $\bbB(R)$.
The iterates $(\u_t)$ of \barons{} are approximate online Newton iterates with respect to the objective functions $(\Phi_t)$ in the sense that for all $t\geq 1$, 
\begin{align}
	\bu_{t+1}\approx \bu_t - \nabla^{-2}\Phi_{t+1}(\u_t)\nabla \Phi_{t+1}(\u_t),\quad \text{for all $t\in[T]$.} \label{eq:approx}
\end{align}
The iterate $\u_{t+1}$ in \barons{} is only an approximation of the right-hand side of \eqref{eq:approx} because \barons{} does not compute the inverse Hessian $\nabla^{-2}\Phi_{t+1}(\u_t)$ exactly at each iteration. Instead, it approximates the inverse Hessian using a Taylor expansion around a neighboring point (see \cref{line:taylor} of \cref{alg:pseudo}).

The motivation behind this approach is that computing the Taylor expansion is significantly cheaper than calculating the exact inverse Hessian. Instead of performing expansions around a fixed point, the algorithm updates the current expansion point $\z_t$ whenever the next iterate $\u_{t+1}$ drifts too far from $\z_t$; see \cref{line:mark}-\cref{line:endmark}. A full inverse Hessian is computed only when the Taylor expansion point is updated. A key insight from \citep{mhammedi2023quasi} is that the iterates of \barons{} are stable enough to ensure that the Taylor expansion point needs to be updated only $O(\sqrt{T})$ times over $T$ rounds, as the next lemma states (the proof can be found in \cref{sec:movement_proof}).
\begin{lemma}[Stability]
	\label{lem:movement}
	Let $\eta,\nu, \Gtilde,R>0$, $T\geq 1$, and $c \in(0,1)$ be given. Consider a call to \cref{alg:pseudo} with input parameters $(T,\eta, \nu, c)$, and let $(\z_t)$ be the Taylor expansion points in \cref{alg:fastexpconcave}. Further, suppose that 
	\begin{itemize}
		\item $\gt_t\in \bbB(\Gtilde)$, for all $t\in [T]$;
		\item $10\Gtilde R  \leq \nu \leq  10 d\Gtilde R T $; and  
		\item $\eta  \leq \frac{1}{5 \Gtilde R}$. 
	\end{itemize}
	 Then, it holds that $\sum_{t=1}^{T-1}  \mathbb{I}\{ \z_{t+1} \neq \z_{t}\} \leq \frac{52}{c} \sqrt{ T \cdot\big(1+\frac{d}{\nu\eta}\ln (1+ \tfrac{T}{d})\big)}.$
\end{lemma}
\paragraph{Computational cost} From a computational perspective, this result is highly promising because it implies that, as long as $\eta$ and $\nu$ are chosen such that $\eta \nu \geq d$, a full Hessian inverse is only required in a $\wtilde{O}(T^{-1/2})$ fraction of the rounds. For the remaining rounds, the computational cost per round is $\wtilde{O}(d^2)$ due to matrix-vector multiplications. As a result, the total computational cost after $T$ rounds is $\wtilde{O}(d^2 T + d^\omega \sqrt{T})$. Therefore, the average per-round computational cost of \barons{} is $\wtilde{O}(d^2)$, assuming $\eta \nu\ge d$ and $T \geq d$\footnote{Here, we are assuming a matrix multiplication exponent $\omega$ satisfying $w\leq  5/2$.}

\paragraph{Feasibility of the iterates} As we show in the analysis, the fact that $(\u_t)$ are approximate online Newton iterates (in the sense of \eqref{eq:approx}) ensures that $(\u_t)$ remain within $\bbB(R)$; it is known (see e.g.~\cite{abernethy2012interior}) that the exact online Newton iterates with respect to $(\Phi_t)$ are guaranteed to stay within $\inte\, \bbB(R)$ due to the self-concordance properties of the regularizer $\Psi$ in the definition of the objectives $(\Phi_t)$.

\subsection{Regret Guarantee} 
The fact that $(\u_t)$ satisfy \eqref{eq:approx} essentially means that it suffices to bound the regret of the Newton iterates. This is advantageous because the (exact) Newton iterates are known to be close to the \ftrl{} iterates $(\w_t)$ with respect to $(\Phi_t)$:
\begin{align}
	\w_t \in \argmin_{\w\in \reals^d} \Phi_t(\w). \label{eq:FTRL}
\end{align}
Due to the curvature of the log-barrier $\Psi$ in the definition of $\Phi_t$ in \eqref{eq:Phi}, a standard \ftrl{} analysis leads to the following regret bound for the iterates $(\w_t)$ (the proof is in \cref{sec:FTRL_proof}).
\begin{lemma}[Regret of \ftrl] 
	\label{lem:FTRL}
	Let $\eta,\nu \in(0,1)$, $R>0$, and $\Gtilde>0$ be such that $\eta \leq \frac{1}{5 \Gtilde R}$ and $\nu \geq 10 \Gtilde R$. Further, let $(\gt_t)\subset \reals^d$ be a sequence of vectors such that $\|\gt_t\|\leq \Gtilde$, for all $t\geq 1$. Then, the \ftrl{} iterates $(\w_t)$ in \eqref{eq:FTRL} satisfy, for all $\w\in \inte\,\bbB(R)$,
	\begin{align}
		\sum_{t=1}^T \inner{\w_t -\w}{\gt_t}  & \leq \sum_{t=1}^T \frac{\eta}{2}\inner{\w_t -\w}{\gt_t}^2 +  \Psi(\w) - \Psi(\bm{0})  + \frac{3d\ln (1 +  T/d)}{\eta}.\nn 
	\end{align}
\end{lemma}
So far, we have outlined that the \barons{} iterates are approximate Newton iterates, which in turn approximate the \ftrl{} iterates. Using these insights, along with the regret bound for \ftrl{} in \cref{lem:FTRL}, we can derive the following regret bound for \barons{} (the proof is in \cref{sec:regretNewton_proof}).
\begin{theorem}[Regret of \barons]
	\label{thm:regretNewton}
	Let $c \in(0,1)$, $T\in \mathbb{N}$, and $\Gtilde >0$ be given and consider a call to \cref{alg:pseudo} with input parameters $(T,\eta,\nu,c)$ such that $\eta \leq  \frac{1}{5 \Gtilde R}$ and $10 \Gtilde R\leq \nu \leq  10 d T\Gtilde R$. If the loss vectors $(\gt_t)$ in \cref{alg:pseudo} satisfy $(\gt_t)\subset \bbB(\Gtilde)$, then the iterates $(\u_t)$ of \cref{alg:pseudo} satisfy $(\u_t)\subset \bbB(R)$ and
	\begin{align}
	\forall \w\in \inte\,\bbB(R),\quad 	\sum_{t=1}^T \left(\inner{\u_t -\w}{\gt_t}- \frac{\eta}{2} \inner{\u_t -\w}{\gt_t}^2\right)\leq    \Gtilde R-\nu \log (1 - \tfrac{\|\w\|^2}{R^2})+ \frac{18 d \log (1+T/d)}{5 \eta}. \label{eq:actual}
	\end{align}
	Further, there is an implementation of \cref{alg:pseudo}, which we display in \cref{alg:fastexpconcave}, with a total computational cost bounded by  \[
	\wtilde O\left(d^2  T \log_c(dT)  + c^{-1}d^\omega \sqrt{\tfrac{d}{\nu \eta} T}\right).\]
\end{theorem}
\subsection{Link to Online Exp-Concave Optimization} The bound in \cref{thm:regretNewton} immediately implies logarithmic regret for online exp-concave optimization. In this setting, $(\gt_t)$ are the subgradients of exp-concave functions $(\ell_t)$, where a function $\ell: \cC \rightarrow \reals^d$ over a convex set $\cC$ is $\alpha$-exp-concave if the mapping $\x \mapsto e^{-\alpha \ell(\x)}$ is concave over $\cC$. It is well known (see e.g., \cite{hazan2007logarithmic}) that for a $\Gtilde$-Lipschitz, $\alpha$-exp-concave function $\ell$, and as long as $\cC \subseteq \bbB(R)$, we have the following for all $\eta \leq \frac{1}{2} \min\big(\frac{1}{4 R \Gtilde}, \alpha\big)$:
\begin{align}
\forall \u,\w\in \cC, \forall \gt\in \partial \ell(\u),\quad \ell(\u)-\ell(\w) \leq \inner{\u-\w}{\gt} - \frac{\eta}{2} \cdot 	\inner{\u-\w}{\gt}^2. 
\end{align} 
This implies that for $\alpha$-exp-concave losses $(\ell_t)$, the corresponding regret $\sum_{t=1}^T(\ell_t(\u_t) - \ell_t(\w))$ can be bounded by the left-hand side of \eqref{eq:actual}, and thus \cref{thm:regretNewton} implies that \barons{} achieves logarithmic regret in this case. Additionally, the term $\sum_{t=1}^T \left(\inner{\u_t -\w}{\gt_t} - \frac{\eta}{2} \inner{\u_t -\w}{\gt_t}^2\right)$ on the left-hand side of \eqref{eq:actual} can itself be interpreted as the regret corresponding to the losses 
\begin{align}
	\ell_t: \x \mapsto  \inner{\x}{\gt_t} + \frac{\eta}{2} \inner{\u_t-\x}{\gt_t}^2,  \label{eq:surrogate}
\end{align}
which are exp-concave for a certain range of $\eta$'s.
\paragraph{Regret improvement over prior work} Compared to \eqref{eq:actual}, the regret bound in \citep{mhammedi2023quasi} includes additional terms like $\wtilde{O}(\Gtilde/\eta)$ and $O(\Gtilde^2)$, with the removal of the latter left as an open problem. As discussed in the next section, in the context of projection-free OCO, $\Gtilde$ is set to $\kappa G$, where $\kappa \coloneqq R/r$, with $r$ and $R$ defined in \cref{ass:setassumption}, and $G$ is the Lipschitz constant of the losses (see \cref{ass:online}). Consequently, having $\kappa$ multiply $1/\eta$ (as in the bound from \citep{mhammedi2023quasi}) would hinder effective tuning of $\eta$ to achieve our desired $\wtilde{O}(\sqrt{d T})$ regret bound for OCO. The improved regret bound for \barons{} in \cref{thm:regretNewton} resolves this issue, as we will see in the next section.

\begin{algorithm}[h]
	\caption{OCO reduction with Gauge projections.}
	\label{alg:projectionfreewrapper}
	\begin{algorithmic}[1]
	\Statex[0] {\bfseries require:} Number of rounds $T$, and an OCO algorithm $\cA$ over $\reals^d$. 
	\State Set $\veps=1/T$ and $\bu_1 = \bm{0}$.
    \State Initialize $\cA$, and set $\u_{1}$ to $\cA$'s first output.
	\For{$t=1,\dots, T$}
	\State Set $(S_t,\bm{s}_t) \gets \gauged{}(\u_t;\K,\veps,r)$. \label{line:Sub} \hfill \algcommentlight{$S_t \approx S_\K(\u_t)$ and $\bs_t \approx \partial S_{\K}(\u_t)$.}
	\State Play $\w_t=  \frac{\u_t}{1+S_t}.$  \label{line:w} \hfill \algcommentlight{$\w_t$ represents an approximate Gauge projection of $\u_t$ onto $\K$.} 
	\State Observe subgradient $\g_t \in \partial f_t(\w_t)$.
	\State Set $ \gt_t = \g_t -  \mathbb{I}\{\inner{\g_t}{\u_t}<0\} \cdot \inner{\g_t}{\w_t} \cdot\bm{s}_t$. \label{line:2} 
	\State Send $\gt_t$ to $\cA$ as the $t$th loss vector.
	\State Set $\u_{t+1} \in \reals^d$ to $\cA$'s $(t+1)$th output given the history $(\u_s,\gt_s)_{s\leq t}$.
	\EndFor
	\end{algorithmic}
	\end{algorithm}	       %
\section{Projection-Free OCO via Exp-Concave Optimization}
\label{sec:oco}
In this section, we demonstrate how, through \cref{alg:projectionfreewrapper}, we can effectively reduce an OCO problem over $\K$ to an online exp-concave problem over a Euclidean ball that contains the feasible set $\K$, allowing us to apply \barons{} from \cref{sec:ons}. Crucially, this reduction only requires Gauge projections (\cref{def:gaugeproj}), which are inexpensive to approximate using a separation oracle; see \cref{sec:gaugedist}. We now provide an overview of our reduction and will elaborate on some of the steps in the sequel.

\subsection{Overview of Reduction} 
Our reduction works as follows:
\begin{enumerate}
\item We have a base algorithm $\cB$ (in this case \cref{alg:projectionfreewrapper}) that outputs feasible points $(\w_t) \subseteq \K$ and observes subgradients $(\g_t \subseteq \partial f_t(\w_t))$;
\item The outputs $(\w_t)$ of $\cB$ are the Gauge projections of the iterates $(\u_t)$ from a subroutine $\cA$; we instantiate $\cA$ as \barons{} over $\bbB(R) \supseteq \K$ in the sequel; \label{item:feasible}
\item Using $(\w_t)$ and $(\g_t)$, the base algorithm $\cB$ constructs surrogate subgradients $(\gt_t)$, which are fed to $\cA$ as loss vectors;
\item The goal is to construct $(\gt_t)$ such that $\|\gt_t\| \leq 2G\kappa$, where $\kappa \coloneqq \frac{R}{r}$ (with $r, R > 0$ as defined in \cref{ass:setassumption}), and 
\begin{align}
  \forall \w\in \K,\quad   \sum_{t=1}^T \left(\inner{\w_t - \w}{\g_t} - \frac{\eta}{2} \inner{\w_t - \w}{\g_t}^2\right) \leq \sum_{t=1}^T \left(\inner{\u_t - \w}{\gt_t} - \frac{\eta}{2} \inner{\u_t - \w}{\gt_t}^2\right) + O(GR); \label{eq:surrogate2}
\end{align} \label{item:surrogate}
\item The sum on the right-hand side of \eqref{eq:surrogate2} is the regret of subroutine $\cA$ with respect to the exp-concave losses $(\ell_t)$ defined in \eqref{eq:surrogate}. By setting $\cA$ to \barons{}, and combining \eqref{eq:surrogate2} with the regret bound of \barons{} in \cref{thm:regretNewton} and the fact that $(\gt_t) \subseteq \bbB(2 \kappa G)$, we essentially obtain
\begin{align}
 \forall \w\in  \K,\quad   \sum_{t=1}^T \inner{\w_t - \w}{\g_t} \leq \frac{\eta}{2} \sum_{t=1}^T \inner{\w_t - \w}{\g_t}^2 + \frac{d}{\eta} \log\left(1 + \frac{T}{d}\right) + O(GR); \label{eq:pretuning}
\end{align} \label{item:ons}
\item Tuning $\eta \propto \min\left(\frac{1}{\kappa GR}, \frac{1}{GR} \sqrt{\frac{d}{T}}\right)$ gives the desired $\wtilde{O}(\sqrt{dT} + \kappa d)$ regret bound.
\end{enumerate}
The key challenge in this reduction is to select surrogate subgradients $(\gt_t)$ that satisfy \eqref{eq:surrogate2} (i.e., \cref{item:surrogate}), while considering that $(\w_t)$ are approximate Gauge projections of $(\u_t)$ (see \cref{item:feasible}). Here, we adopt a similar choice of surrogate subgradients as in \citep{mhammedi2022efficient}.

\paragraph{Choice of surrogate subgradients}
At round $t\geq 1$, given $\u_t$, $\w_t$, and $\g_t$, \cref{alg:projectionfreewrapper} sets the surrogate subgradient as $\gt_t\approx \gproj_t$, where 
\begin{align}
\gproj_t \in \g_t - \mathbb{I}\{\inner{\g_t}{\u_t}<0\} \cdot \inner{\g_t}{\w_t} \cdot   \partial	S_\K(\u_t), \label{eq:sur}
\end{align} 
and $S_{\K}(\u) \coloneqq \min_{\x\in \K}\gamma_\K(\u - \x)$ is the Gauge distance function. \cref{alg:projectionfreewrapper} computes an approximate subgradient $\bs_t$ of $S_\K$ at $\u_t$ using the $\gauged$ subroutine. By \cref{lem:approxgrad}, we have that $S_\K(\w) \geq S_\K(\u_t) + \inner{\w - \u_t}{\bs_t} - \veps$ for all $\w \in \reals^d$ ($\veps$ is set to $1/T$ in \cref{alg:projectionfreewrapper}), which essentially shows that $\bs_t$ is an approximate subgradient of $S_\K$ at $\u_t$. This, in turn, implies that $\gt_t \approx \gproj_t$ by \cref{line:2} of \cref{alg:projectionfreewrapper} and \eqref{eq:sur}. Moreover, as noted in \cref{lem:approxgrad}, computing $\gt_t$ is computationally inexpensive, requiring only $\wtilde{O}(1)$ calls to the separation oracle.

\paragraph{Approximate Gauge projections} To ensure feasible iterates, \cref{alg:projectionfreewrapper} sets $(\w_t)$ as approximate Gauge projections of the outputs $(\u_t)$ from $\cA$ onto $\K$; that is, $\w_t \approx \wproj_t$, where $\wproj_t \in \argmin_{\x \in \K} \gamma_{\K}(\u_t - \x)$ for all $t \geq 1$. By \cref{lem:gauge}, the exact Gauge projection points $(\wproj_t)$ satisfy \begin{align}
	\wproj_t & = \frac{\u_t}{1+S_\K(\u_t)},\label{eq:project}
\end{align}  
since $S_\K(\u_t)=\max(0,\gamma_\K(\u_t)-1)$; see \eqref{eq:dist}. In \cref{alg:projectionfreewrapper}, we approximate $\gamma_\K(\u_t)$ using the $\gauged$ subroutine (\cref{alg:gauge}). By \cref{lem:approxgrad}, we have that $S_t$ in \cref{alg:projectionfreewrapper} satisfies $S_\K(\u_t) \leq S_t \leq S_\K(\u_t) + \veps$. Thus, by \cref{line:w} of \cref{alg:projectionfreewrapper} and \eqref{eq:project}, we indeed have that $\w_t \approx \wproj_t$ and $\w_t \in \K$ for all $t \geq 1$ (the details are in the proof of \cref{lem:inst} in \cref{sec:instproof}). As noted in \cref{lem:approxgrad}, a call to $\gauged$ requires only $\wtilde{O}(1)$ calls to the separation oracle $\Sep_\K$. Therefore, \cref{alg:projectionfreewrapper} ensures feasibility with only a logarithmic number of oracle calls.

\subsection{Guarantees of Reduction}
The specific choice of surrogate subgradients in \eqref{eq:sur} ensures that $\inner{\g_t}{\wproj_t - \w} \leq \inner{\gproj_t}{\u_t - \w}$ for all $\w \in \K$; this follows from the analysis in \citep{mhammedi2022efficient}. Taking into account the approximation errors $\w_t \approx \wproj_t$ and $\gt_t \approx \gproj_t$, we obtain the following result (the proof is in \cref{sec:instproof}).%
\begin{lemma}[Key reduction result]
	\label{lem:inst}	
	Let $T\geq 1$ be given, and suppose that \cref{ass:setassumption} and \cref{ass:online} hold. Further, let $(\w_t)$, $(\u_t)$, $(\g_t)$, and $(\gt_t)$ be as in \cref{alg:projectionfreewrapper} with input $T$. If the iterates $(\u_t)$ of the subroutine $\cA$ satisfy $(\u_t)\subset \bbB(R)$, then we have $\|\gt_t\|\leq 2\kappa \cdot G$, where $\kappa \coloneqq \frac{R}{r}$, and for all $t\in [T]$:
	\begin{align}
		\w_t\in \K 	 \qquad \text{and} \qquad \forall \w\in \K,\quad 
		\inner{\g_t}{\w_t- \w} 
		\leq 	\inner{\gt_t}{\u_t- \w}+ \frac{2 GR}{T}   .  \label{eq:monotone_pre}
	\end{align}
\end{lemma}
By summing \eqref{eq:monotone_pre} over $t=1,\dots, T$, we get that 
\begin{align}
\forall \w\in \K,\quad \sum_{t=1}^T \inner{\g_t}{\w_t- \w} \leq 	\sum_{t=1}^T \inner{\gt_t}{\u_t- \w} + 2 G R. \label{eq:transfer}
\end{align}
\begin{remark}[Recovering existing regret bounds]
	\label{rem:recover}
	At this point, we can already recover the $O(\kappa \sqrt{T})$ regret bound of existing separation-based projection-free algorithms \citep{mhammedi2022efficient,garber2022new}. The right-hand side of \eqref{eq:transfer} represents the regret of the subroutine $\cA$ (with respect to the losses $(\w \mapsto \inner{\gt_t}{\w})$). If we instantiate $\cA$ as projected gradient descent over the ball $\bbB(R)$ (where Euclidean projections cost $O(d)$) and use the fact that $\|\gt_t\| \leq 2\kappa G$ for all $t \geq 1$ (by \cref{lem:inst}), we obtain $\sum_{t=1}^T \inner{\gt_t}{\u_t - \w} \leq O(\kappa \sqrt{T})$. Combining this with \eqref{eq:transfer} results in an overall $O(\kappa \sqrt{T})$ regret bound  for \cref{alg:pseudo}, which importantly makes only $\wtilde{O}(1)$ calls to a separation oracle per round. Note that \barons{} was not needed for this part.
\end{remark}

Returning to our reduction, we note that the inequality in \eqref{eq:transfer} is similar but does not exactly match the inequality we seek in \cref{item:surrogate}, as the terms $\frac{\eta}{2}\sum_{t=1}^T \inner{\g_t}{\w_t - \w}^2$ and $\frac{\eta}{2}\sum_{t=1}^T \inner{\gt_t}{\u_t - \w}^2$ are missing from \eqref{eq:transfer}. However, we can still derive the inequality in \cref{item:surrogate}, starting from \cref{lem:inst}. The full details are in the proof of \cref{lem:meta} in \cref{sec:surro}, but here we provide a sketch. 

Suppose that $(\u_t) \subset \bbB(R)$ (which holds if $\cA$ is set to \barons{} by \cref{thm:regretNewton}). Since the function $x \mapsto x - \eta x^2 / 2$ is non-decreasing for $x \leq 1/2$, by choosing $\eta \leq \frac{1}{10\kappa G R}$, setting $x$ to $\inner{\g_t}{\w_t - \w}$ and $\inner{\gt_t}{\u_t - \w} + \frac{2 GR}{T}$, and applying \cref{lem:inst}, we get that for any $t\in[T]$ and $\w\in \K$:
\begin{align}
\inner{\g_t}{\w_t- \w} - \frac{\eta}{2} \inner{\g_t}{\w_t - \w}^2 &\leq 	\inner{\gt_t}{\u_t- \w} + \frac{2GR}{T}-\frac{\eta}{2} \inner{\gt_t}{\u_t- \w}^2 - \frac{2\eta GT}{T} \inner{\gt_t}{\bu_t-\w} -\frac{2\eta G^2R^2}{T^2},\nn \\
	& \leq \inner{\gt_t}{\u_t- \w}-\frac{\eta}{2} \inner{\gt_t}{\u_t- \w}^2+ \frac{3 G R}{T}, \label{eq:transfer2}
	\end{align}
	where the last inequality follows by the facts that for all $t\in[T]$ and $\w\in \K$:
	\begin{itemize} 
		\item $|\inner{\gt_t}{\u_t-\w}|\leq \|\gt_t\|\cdot \|\u_t-\w\|$ by Cauchy Schwarz;
		\item $\|\gt_t\|\leq 2\kappa G$ by \cref{lem:inst}; 
		\item $\|\u_t-\w\|\leq 2 R$, since $\u_t,\w\in \bbB(R)$; and 
		\item  $\eta \leq \frac{1}{10\kappa GR}$.
	 \end{itemize}
	 Summing \eqref{eq:transfer2} over $t = 1, \dots, T$ yields \eqref{eq:surrogate2}. As noted in \cref{item:ons}, the sum on the right-hand side of \eqref{eq:surrogate} represents the regret of the subroutine $\cA$ with respect to the exp-concave losses $(\ell_t: \x \mapsto \inner{\gt_t}{\x} + \frac{\eta}{2} \inner{\gt_t}{\u_t - \x}^2)$. Thus, by setting $\cA$ to \barons{} and using the regret bound of \barons{} in \cref{thm:regretNewton}, we recover the regret bound in \eqref{eq:pretuning}. We now state this result (the proof can be found in \cref{sec:surro}).

\begin{proposition}
	\label{lem:meta}	
	Let $c\in(0,1)$ and $T\geq 1$ be given. Suppose that \cref{ass:setassumption} and \cref{ass:online} hold. Consider a call to \cref{alg:projectionfreewrapper} with input $T$ and where the subroutine $\cA$ is an instance of \barons{} (\cref{alg:pseudo}) with input parameters $(T,\eta,\nu,c)$ satisfying $\eta \leq \frac{1}{10 \kappa R G}$, and $20 \kappa GR\leq \nu \leq 20 d\kappa G R$, where $\kappa$ is as in \cref{lem:inst}. Then, the sequences $(\w_t)$ and $(\g_t)$ in \cref{alg:projectionfreewrapper} satisfy:
\begin{align}
\forall \w\in \inte\, \K,  \quad 
	\sum_{t=1}^T \inner{\g_t}{\w_t -\w}  \leq  \frac{\eta}{2} \sum_{t=1}^T\inner{\g_t}{\w_t -\w}^2 +  5\kappa G R  - \nu \log (1 - \tfrac{\|\w\|^2}{R^2})+ \frac{4 d \log (1+\frac{T}{d})}{\eta}.   
\end{align}
\end{proposition}
To get our main $\wtilde{O}(\sqrt{d T})$ regret bound for \cref{alg:projectionfreewrapper}, we instantiate \cref{lem:meta} with the parameters:
\begin{align}
	c=\tfrac{1}{2},\quad 
	\eta = \tfrac{1}{GR} \cdot \min \left(\tfrac{1}{10 \kappa}, \sqrt{\tfrac{2 d \log (1 + \frac{T}{d})}{ T}}\right), \quad \text{and} \quad \nu = G R \cdot \max \left(20 \kappa d, \sqrt{\tfrac{dT}{\log (1+ \frac{T}{d})} }\right),
	\label{eq:parameterchoice}
\end{align} where $\kappa$ is as in \cref{lem:inst}.
\begin{theorem}[Main guarantee]
\label{cor:main}
Let $T\geq 1$ be given, and suppose that \cref{ass:setassumption} and \cref{ass:online} hold. Consider a call to \cref{alg:projectionfreewrapper} with input $T$ and where the subroutine $\cA$ is an instance of \barons{} (\cref{alg:pseudo}) with input parameters $(T,\eta,\nu,c)$ as in \eqref{eq:parameterchoice}.
Then, $(\w_t)$ and $(\g_t)$ in \cref{alg:projectionfreewrapper} satisfy:
\begin{align}
\forall \w\in \K,  \quad 
\sum_{t=1}^T \inner{\g_t}{\w_t -\w}  \leq  5 GR \sqrt{2 d T \log (1+ \tfrac{T}{d})}  +  66  G R \kappa d \log(1 + \tfrac{T}{d}).   \label{eq:monotone_new}
\end{align}
Furthermore, the computation cost of the instance of \cref{alg:projectionfreewrapper} under consideration is bounded by 
\begin{align}
 \wtilde{O} \left(C_{\sep}(\K) \cdot  T +  d^2\cdot T + d^\omega \sqrt{T}\right), \label{eq:cost}
\end{align}
where $C_\sep(\K)$ is the cost of one call to a separation oracle for the set $\K$.
\end{theorem}
The full proof of \cref{cor:main} is deferred to \cref{sec:proofoco}. Here, we sketch why the computational cost of the instance of \cref{alg:projectionfreewrapper} in \cref{cor:main} is bounded by \eqref{eq:cost}.

\begin{proof}[Proof sketch of \eqref{eq:cost}] By \cref{thm:regretNewton}, the computational cost of the \barons{} subroutine within \cref{alg:projectionfreewrapper} is bounded by $\wtilde{O}\left( d^2 T +  d^{\omega} \sqrt{\tfrac{dT}{\nu \eta}} \right)$. Given the choice of $\eta$ and $\nu$ in \eqref{eq:parameterchoice}, we have $\eta \nu \geq d$, implying that the computational cost of the \barons{} subroutine is bounded by
\[
\wtilde{O}\left(d^2 T +  d^{\omega}\sqrt{T}\right).
\]

In addition to the cost of the \barons{} subroutine, \cref{alg:projectionfreewrapper} incurs $O(d + C_{\gauged}(\cK))$ per round, where $C_{\gauged}(\cK)$ represents the cost of calling the $\gauged$ subroutine to approximate the gauge distance $S_\K$ and its subgradient. By \cref{lem:approxgrad}, we have
\[
C_{\gauged}(\cK) \leq \wtilde{O}(1) \cdot C_{\sep}(\cK).
\]
This implies the desired computational cost in \eqref{eq:cost}.
\end{proof}
\subsection{Computational Cost and Additional Results}
If we take the matrix exponent $\omega$ to be $\omega \leq 5/2$, the average per-round computational cost of our approach is bounded by $\wtilde{O}(C_\sep(\K) + d^2)$ as long as $T\geq d$. In contrast, existing separation-based projection-free algorithms, such as those in \citep{mhammedi2022efficient,garber2022new,lu2023projection}, have a per-round computational cost bounded by $\wtilde{O}(C_\sep(\K) + d)$ but guarantee a regret bound of $O(\kappa \sqrt{T})$. However, as discussed in the introduction, $\kappa$ can be arbitrarily large for ill-conditioned sets. As shown in \citep{flaxman2005online}, one can apply an affine transformation to $\K$ (e.g., to put it in isotropic position) to ensure that $\kappa$ is at most $d$. Doing so, however, incurs an additional $O(d^2)$ computational cost per round, as it requires multiplying the incoming subgradients by the matrix corresponding to the affine transformation at each round. This adjustment brings the overall computational cost in line with the cost of \cref{alg:projectionfreewrapper} in \eqref{eq:cost}.

Thus, one can think of the $O(d^2)$ cost of our approach as the ``price'' for adapting to the ill-conditioning of the set without needing to compute an affine transformation, while still ensuring a $\wtilde{O}(\sqrt{dT})$ regret bound—improving on the worst-case regret bounds of previous separation-based projection-free algorithms by a factor of $\sqrt{d}$ (because $\kappa \sqrt{T}$ can be as large as $d\sqrt{T}$ even after pre-processing). Finally, we note that the $\wtilde{O}(d^2)$ in our final algorithm cost comes from matrix-vector multiplications, which are easily parallelizable.

\paragraph{Best-of-both-worlds bound} Using standard aggregation techniques, such as \texttt{Hedge}{} \citep{freund1997decision}, we can achieve a “best-of-both-worlds” regret bound of $\wtilde{O}((\kappa \wedge\sqrt{d})\cdot\sqrt{T})$, which is particularly beneficial when $\kappa$ is small. 
 
\paragraph{Adaptive regret bounds}
We note that, since \cref{lem:inst} allows us to bound the \emph{instantaneous} regret of \cref{alg:projectionfreewrapper} by the instantaneous regret of the subroutine $\cA$ (see also \eqref{eq:surrogate2}), any special guarantees that $\cA$ possesses readily transfer to \cref{alg:projectionfreewrapper}. For example, if $\cA$ has an adaptive guarantee \citep{hazan2009efficient}, \cref{alg:projectionfreewrapper} will inherit that guarantee as well.

\section{Application to Offline and Stochastic Optimization}
\label{sec:stochastic}
In this section, we leverage the results from the previous sections to achieve a state-of-the-art convergence rate for projection-free stochastic convex optimization. We begin by presenting our results for stochastic convex optimization, then specialize them to the offline convex optimization setting for finding a near-optimal point. To proceed, we now state our assumption for the stochastic optimization setting.

\begin{assumption}
\label{assum:stoch}
There is a function $f:\K \rightarrow \reals$ and parameters $\sigma\geq 0$ and $G>0$ such that the loss vector $\g_t$ that the algorithm receives at round $t\geq 1$ is of the form $\g_t = \gb_t + \bxi_t$, where 
\begin{itemize}
\item For all $t\geq 1$, $\gb_t\in \partial f(\w_t)$, where $\w_t$ is the output of the algorithm at round $t$;  
\item $(\bxi_t)\subset \reals^d$ are i.i.d.~noise vectors such that $\E[\bxi_t]=\bm{0}$ and $\E[\|\bxi_t\|^2]\leq \sigma^2$, for all $t\geq 1$; and
\item For all $t\geq 1$, $\|\gb_t\|\leq G$.
\end{itemize}
\end{assumption}
\subsection{Convergence Rate in Stochastic Convex Optimization}
Under \cref{assum:stoch}, we now state our main guarantee for the stochastic convex optimization setting. As in the OCO setting, we use \cref{alg:projectionfreewrapper} with the subroutine $\cA$ set as \barons{}. Here, we set the parameters of \barons{} as
\begin{align}c=\tfrac{1}{2},\quad 
\eta = \tfrac{1}{R} \cdot \min \left(\tfrac{1}{10 G\kappa}, \tfrac{1}{\sigma}\sqrt{\tfrac{2 d \log (1 + \frac{T}{d})}{ T}}\right), \quad \text{and} \quad \nu =  R \cdot \max \left(20 G\kappa d, \sigma \sqrt{\tfrac{d T}{\log (1+ \frac{T}{d})} }\right),
  \label{eq:params2}
\end{align} 
where $\kappa\coloneqq R/r$ and $r,R>0$ are as in \cref{ass:setassumption}. 
\begin{theorem}
\label{prop:stoch}
Let $T>0$ be given. Suppose that \cref{ass:setassumption} and \cref{assum:stoch} (with $\sigma, G\geq 0$) hold and consider a call to \cref{alg:projectionfreewrapper} with input $T$, where the subroutine $\cA$ is an instance of \barons{} (\cref{alg:pseudo}) with input parameters $(T,\eta,\nu,c)$ as in \eqref{eq:params2}. Then, we have
\begin{align} 
\E[f(\widehat{\w}_T)] - \inf_{\w\in \K}f(\w)  \leq  16 R \sigma \sqrt{\frac{d \log (1+ \tfrac{T}{d})}{T}}  +  \frac{74 G R\kappa d\log(1 +\frac{T}{d})}{T}, \label{eq:stoch}
\end{align}
where $\widehat{\w}_T \coloneqq \frac{1}{T} \sum_{t=1}^T\w_t$ and $(\w_t)$ are the iterates of \cref{alg:projectionfreewrapper}. The computational cost is bounded by \eqref{eq:cost}.
\end{theorem}
The proof of the theorem, which follows from an application of \cref{lem:meta}, is in \cref{sec:stoch}. Extending the result in \cref{prop:stoch} to a high-probability guarantee is possible using martingale concentration bounds. 

\subsection{Computational Cost of Finding a Near-Optimal Point}
We now consider the computational cost for finding a near-optimal point in offline convex optimization; that is, when $\sigma =0$.
In this case, from \eqref{eq:stoch}, if we set \[T = \mathfrak{c} \cdot  \frac{G R\kappa d }{\veps},\] with $\mathfrak{c} = \polylog(d,1/\veps)$ sufficiently large, we get that 
\begin{align}
    f(\what\w_T)- \inf_{\w\in \K}f(\w) \leq \veps,
\end{align}
where $\widehat{\w}_T \coloneqq \frac{1}{T} \sum_{t=1}^T\w_t$ and $(\w_t)$ are the iterates of \cref{alg:projectionfreewrapper}. Thus, $\widehat{\w}_T$ represents an $\veps$-optimal point for the objective function $f$. Now, by \cref{prop:stoch}, the computational cost of the instance of \cref{alg:projectionfreewrapper} in \cref{prop:stoch} is bounded by \eqref{eq:cost}. Instantiating \eqref{eq:cost} with the choice of $T = \mathfrak{c} \cdot \frac{G R \kappa d}{\veps}$, which implies that $d^2T \gg \Omega(d^{\omega} \sqrt{T})$ (as long as $\omega \leq 5/2$), the cost of finding an $\veps$-optimal point in offline convex optimization using our approach is bounded by
\begin{align}
    \frac{\kappa d}{\veps} \cdot (C_\sep(\K) + d^2),
\end{align}
where we recall that the cost $O(d^2)$ is coming from matrix-vector multiplications, which is parallelizable.

\section{Conclusion and Future Work}
\label{sec:conclusion}
We presented a new separation-based algorithm for OCO that achieves a regret bound of $\wtilde{O}(\sqrt{d T} + \kappa d)$ while making only $\wtilde{O}(1)$ calls to a separation oracle per round. Existing separation-based algorithms guarantee a regret bound of $O(\kappa \sqrt{T})$. The asphericity factor $\kappa$ can be arbitrarily large for ill-conditioned feasible sets and can reach as high as $d$, even after applying the best affine transformation to the feasible set. In this case, the main term in our regret bound, $\wtilde{O}(\sqrt{dT})$, improves over state-of-the-art bounds by a factor of $\sqrt{d}$ without requiring any preconditioning.

While we have successfully eliminated some unwanted factors from existing bounds, an open question remains: is there a projection-free algorithm (whether separation or linear optimization-based) that can achieve the optimal, dimension-free $O(\sqrt{T})$ regret while making only $\wtilde{O}(1)$ oracle calls per round? A less ambitious but still important goal would be to achieve our $\wtilde{O}(\sqrt{dT})$ regret bound without the need for matrix multiplications or storing a matrix (as required by \barons), which incurs a memory cost of $d^2$.

	\newpage
	\bibliography{../refs3.bib}
	
	\newpage

	\appendix

\clearpage
\addcontentsline{toc}{section}{Appendix} %
\section{Organization of the Appendix}
This appendix is organized as follows:

\begin{itemize}
	\item In \cref{sec:fullv}, we present the full version of the \barons{} algorithm.
	\item \cref{sec:self-concordant} provides background on self-concordant functions, highlighting key properties used in our analysis.
	\item  In \cref{sec:ons_appendix}, we present the proof for the regret guarantee of \barons{} in \cref{thm:regretNewton}.
	\item In \cref{sec:ons_appendix}, we present the proof of our main OCO result in \cref{cor:main}.
	\item In \cref{sec:stoch}, we provide the proof of \cref{prop:stoch} for the convergence rate of our algorithm in the stochastic convex optimization setting.
	\item Finally, \cref{sec:fullproof} includes  the proof of \cref{lem:gauge} on the approximate computation of the Gauge distance and its subgradients.
\end{itemize}

\section{Full Version of \barons}
\label{sec:fullv}
\begin{algorithm}
		\caption{Efficient implementation of \barons{} (\cref{alg:pseudo}). The current algorithm is technically equivalent to \cref{alg:pseudo} in that both algorithms produce the same outputs.} 
		\label{alg:fastexpconcave}
		\begin{algorithmic}[1]
			\Statex[0]{\bf input} Parameters $\eta,\nu, c >0$. 
			\State  \multiline{Set $\z_1\gets\bm{0}$, $\u_1 \gets \bm{0}$, $V_0 \gets 0$, $\Sigma'_0 \gets  \frac{1}{2\nu}I$, $S_0\gets \bm{0}$, $G_0\gets \bm{0}$, and $m=\mathfrak{c}\log_c(dT)$, with $\mathfrak{c}$ a sufficiently large universal constant.}
			\For{$t=1,\dots, T$}
			\State Play $\u_t$ and observe $\g_t \in \partial \ell_t(\u_t)$.
			\State  Set $G_t \gets G_{t-1}+ \g_t$, $S_t \gets S_{t-1} +\g_t \g_t^\top \u_t$, and $V_t \gets V_{t-1}+\g_t\g_t^\top$. \label{line:dil}
			\State Set $\bnabla_t\gets  \frac{2\nu \u_{t}}{1-\|\u_t\|^2}  +\eta V_t\u_t- \eta S_t+ G_t$.\label{line:purge}\hfill \algcommentlight{$\bnabla_t = \bnabla \Phi_{t+1}(\u_t)$}
			\State \label{line:class}Set $\Sigma'_t \gets   \Sigma'_{t-1}  -  \frac{\eta \Sigma'_{t-1} \g_t \g_t^\top \Sigma'_{t-1}}{1+\eta \g_t^\top \Sigma'_{t-1} \g_t}$ and $\Sigma_t \gets \Sigma'_t - \frac{4\nu   \Sigma'_{t} \u_t \u_t^\top \Sigma'_{t}}{(R^2-\|\u_t\|^2)^2+4 \nu  \u_t^\top \Sigma'_{t} \u_t}$. 
            \Statex[1] \algcommentbiglight{Computing the Taylor expansion}
			\State Set $\bdelta_{t}\gets \Sigma_t \bnabla_t$ and $\tilde\bdelta_{t}\gets \Sigma_t \bdelta_t$. \label{line:set}
			\For{$k=1,\dots,m$}
			\State Set $\tilde\bdelta_{t}\leftarrow   \left(\frac{2\nu }{R^2-\|\z_t\|^2} - \frac{2\nu  }{R^2-\|\u_t\|^2}\right)  \Sigma_t  \tilde\bdelta_{t}$.
			\State Set $\bdelta_{t} \leftarrow \bdelta_t+ \tilde\bdelta_{t}$.
			\EndFor
            \Statex[1] \algcommentbiglight{Perform approximate Newton step}
			\State  Set $\u_{t+1}\gets\u_t -  \bdelta_t$.\hfill \algcommentlight{$\u_{t+1} \approx \u_{t}- \bnabla^{-2} \Phi_{t+1}(\u_t) \bnabla\Phi_{t+1}(\u_t).$} \label{line:update} %
            \Statex[1] \algcommentbiglight{Check if the Taylor expansion point needs to be updated}
			\If{$|\|\u_{t+1}\|^2-\|\z_{t}\|^2| \leq  c \cdot (R^2-\|\z_t\|^2)$} \label{eq:lineeight}
			\State Set $\z_{t+1}\gets\z_{t}$.
			\Else 
			\State Set $\z_{t+1}\gets\u_{t+1}$.
			\State \label{line:fbi} Set $\Sigma'_t = \left(\frac{2\nu  I}{R^2-\|\u_{t+1}\|^2}  +{\eta} V_t\right)^{-1}$. \hfill \algcommentlight{$\Sigma'_t\gets \left(\bnabla^2 \Phi_{t+1}(\u_{t+1})- \frac{4\nu \u_{t+1} \u_{t+1}^\top}{(R^2-\|\u_{t+1}\|^2)^2} \right)^{-1}$}  \label{line:endmark2}
			\EndIf 
			\EndFor
		\end{algorithmic}
	\end{algorithm}

\section{Self-Concordant Functions}
	\label{sec:self-concordant}
In this section, we introduce the concept of self-concordant functions and outline several key properties that play an important role in our proofs (the results presented in this section are taken from \citep{mhammedi2023quasi}). We begin by defining a self-concordant function. For the rest of this section, let $\cC$ represent a convex, compact set with a non-empty interior, denoted by $\inte\, \cC$. For a function that is twice differentiable [or thrice differentiable], we denote its Hessian as $\nabla^2 f(\u)$ [and its third derivative tensor as $\nabla^3 f(\u)$] at $\u$.
	\begin{definition}
		A convex function $f\colon \inte\, \cC \rightarrow \reals$ is called \emph{self-concordant} with constant $M_f\geq 0$, if $f$ is $C^3$ and satisfies 
\begin{itemize}
	\item 		 $f(\x_k)\to +\infty$ for $\x_k \to \x\in \partial \cC$; and
	\item For all $\x\in  \inte\, \cC$ and $\u \in \reals^d$:
		\begin{align}
 |\nabla^3f(\x)[\u,\u,\u]| \leq 2 M_f \|\u\|^3_{\nabla^2 f(\x)}.
		\end{align}
	\end{itemize}
	\end{definition} 
	By definition, if $f$ is self-concordant with a constant $M_f \geq 0$, it remains self-concordant for any constant $M \geq M_f$. For a self-concordant function $f$ and a point $\x \in \dom f$, the quantity $\lambda(\x, f) \coloneqq \|\nabla f(\x)\|_{\nabla^{-2}f(\x)}$, referred to as the \emph{Newton decrement}, plays a key role in our proofs. The following two lemmas summarize properties of the Newton decrement and the Hessians of self-concordant functions, which will be utilized frequently in the proofs of \barons{} (see, for example, \cite{nemirovski2008interior,nesterov2018lectures}).
	\begin{lemma}
		\label{lem:properties}
		Let $f\colon \inte\, \cC\rightarrow \reals$ be a self-concordant function with constant $M_f\geq 1$. Further, let $\x\in \inte\, \cC$ and $\x_f\in \argmin_{\x\in \cC} f(\x)$. Then, 
        \begin{itemize}
            \item Whenever $\lambda(\x,f)<1/M_f$, we have 
		\begin{align}
			\|\x -\x_f\|_{\nabla^2 f(\x_f)} 	\vee 	\|\x -\x_f\|_{\nabla^2 f(\x)} \leq {\lambda(\x,f)}/({1-M_f \lambda (\x,f)});  \nn
		\end{align}
        \item For any $M\geq M_f$, the \emph{Newton step} $\x_+\coloneqq \x - \nabla^{-2}f(\x)\nabla f(\x)$ satisfies $\x_+\in \inte\, \cC$ and 
		\[\lambda(\x_+,f)\leq M \lambda(\x,f)^2/ ( 1 - M \lambda(\x,f))^2.\]
    \end{itemize}
	\end{lemma}
	\begin{lemma}
		\label{lem:hessians}
		Consider a self-concordant function $f\colon \inte, \cC \rightarrow \reals$ with constant $M_f$ and a point $\x \in \inte, \cC$. For any $\y$ such that $r \coloneqq \|\y - \x\|_{\nabla^2 f(\x)} < 1/M_f$, the following holds:
		\begin{align}
			(1-M_f r)^{2} \nabla^2 f(\y) \preceq \nabla^2 f(\x) \preceq (1-M_f r)^{-2}  \nabla^2 f(\y).\nn 
		\end{align}
	\end{lemma}
	The following result from \cite[Theorem 5.1.5]{nesterov2018lectures} will be helpful in showing that the iterates of our algorithms consistently remain within the feasible set.
	\begin{lemma}
		\label{lem:deakin}
		Let $f\colon \inte\, \cC\rightarrow \reals$ be a self-concordant function with constant $M_f\geq 1$ and $\x \in \inte\, \cC$. Then, $\cE_{\x} \coloneqq \{\w \in \reals^d\colon \|\w-\x\|_{\nabla^{2}f(\x)}<1/M_f \}\subseteq \inte\, \cC$. Furthermore, for all $\w\in \cE_{\x}$, we have $$\|\w-\x\|_{\nabla^2 f(\w)} \leq \frac{\|\w- \x\|_{\nabla^2 f(\x)}}{1-M_f \|\w- \x\|_{\nabla^2 f(\x)}}.$$
	\end{lemma}
	Finally, we also need the following result due to \cite{mhammedi2022newton}.
	\begin{lemma}
		\label{lem:inter0} 
		Let $f\colon \inte\, \cC \rightarrow \reals$ be a self-concordant function with constant $M_{f}>0$. Then, for any $\x, \y \in \inte\, \cC$ such that $r\coloneqq \|\x-\y\|_{\nabla^2 f(\x)}<1/M_{f}$, we have 
		\begin{align}
			\|\nabla f(\x) - \nabla f(\y)\|^2_{\nabla^{-2}f(\x)}  \leq \frac{1}{(1-M_{f} r )^{2}} \|\y- \x\|^2_{\nabla^{2}f(\x)}.\nn
		\end{align}
	\end{lemma}
 
\section{ONS Analysis: Proof of \cref{thm:regretNewton}}
\label{sec:ons_appendix} 

This appendix provides a proof of \cref{thm:regretNewton}. Each subsection provides results and proofs of intermediate results outlined in \cref{sec:ons}.
\subsection{Taylor Expansion of Inverse Hessians}
In this section, we prove that the Taylor expansions used in \cref{alg:pseudo} indeed approximate the inverse Hessians $(\nabla^{-2}\Phi_{t+1}(\u_t))$. This result we state next is a slight modification of a result in \cite{mhammedi2023quasi}.
\begin{lemma} 
    \label{lem:hessian}
    Let $\eta,\nu>0$, $c\in(0,1)$, and $T\geq 1$ be given. Further, let $(T,\gamma_t)$ and $(\Sigma_t)$ be as in \cref{alg:pseudo} with parameters $(\eta, \nu, c)$. Then, for $t\in[T]$ such that $\u_t\in \inte\,\bbB(R)$ and for any $m\geq 1$, we have
    \begin{align}
        \left\| \nabla^{-2}\Phi_{t+1}(\u_t) - \sum_{k=1}^{m+1} \gamma_t^{k-1} \Sigma_t^{k} \right\|	\leq  \frac{R^2 c^m}{2\nu \cdot(1-c)}. \nn 
    \end{align} 	
\end{lemma}

\begin{proof}
    Fix $m\geq 1$ and let $\alpha_t \coloneqq \frac{ \|\u_t\|^2-\|\z_t\|^2}{R^2-\|\u_t\|^2}$. We have 
    \begin{align}
        \alpha_t = \frac{R^2-\|\z_t\|^2}{R^2-\|\u_t\|^2}- 1= -\frac{R^2-\|\z_t\|^2}{2\nu } \gamma_t,\nn
    \end{align}
    where we recall that $\gamma_t =\frac{2 \nu}{R^2-\|\z_t\|^2} - \frac{2\nu}{R^2-\|\u_t\|^2}$.
    Note that $\Sigma_t$ in \cref{alg:pseudo} satisfies 
    \begin{align}
        \Sigma_t^{-1} &=  \frac{2 \nu  I}{R^2-\|\z_t\|^2} + \frac{4\nu \u_t \u_t^\top}{(R^2-\|\u_t\|^2)^2}  + \eta \sum_{s=1}^{t-1} \gt_s \gt_s^\top,\label{eq:slide} \\
        &  = \nabla^2 \Phi_{t+1}(\u_t) - \frac{2\nu  I}{R^2-\|\u_t\|^2}+\frac{2\nu  I}{R^2-\|\z_t\|^2},\nn \\
        & = \nabla^2 \Phi_{t+1}(\u_t) - \frac{2\nu  I}{R^2-\|\z_t\|^2}\cdot  \left( \frac{R^2-\|\z_t\|^2}{R^2-\|\u_t\|^2}-1\right),\nn \\
        & = \nabla^2 \Phi_{t+1}(\u_t) -  \frac{2\nu  \alpha_tI}{R^2-\|\z_t\|^2}. \nn 
    \end{align}
    Therefore, if we let $U_t \coloneqq (R^2-\|\z_t\|^2)H^{-1}_t/(2\nu )$, we have 
    \begin{align}
        \nabla^{-2}\Phi_{t+1}(\u_t) & =  \left( \frac{2\nu  \alpha_t I}{R^2-\|\z_t\|^2} + H^{-1}_t\right)^{-1},\nn\\
        & = \frac{R^2-\|\z_t\|^2}{2\nu } \left( \alpha_t I +  \frac{R^2-\|\z_t\|^2}{2\nu }H^{-1}_t\right)^{-1},\nn \\
        & = \frac{R^2-\|\z_t\|^2}{2\nu } ( \alpha_t I + U_t)^{-1},\nn \\
        & =    \frac{R^2-\|\z_t\|^2}{2\nu }  U_t^{-1}  (I+\alpha_t U_t^{-1})^{-1}. \label{eq:cry} 
    \end{align}
    Now, by \eqref{eq:slide}, we have $U_t \succeq I$ and so $\|U^{-1}_t\| \leq 1$. Using this and that $|\alpha_t| \leq c <1$ (this is an invariant of \cref{alg:pseudo}---see \cref{line:mark} of \cref{alg:pseudo}), we have 
    \begin{align}
        (1+\alpha_t U^{-1}_t)^{-1} =	 \sum_{k=0}^{\infty}(-\alpha_t)^k U_t^{-k}, \quad \text{and} \quad \left\| (1+\alpha_t U_t)^{-1} - \sum_{k=0}^{m} (-\alpha_t)^k U_t^{-k} \right\| \leq \frac{c^m}{1-c}. \nn 
    \end{align}
    Therefore, by \eqref{eq:cry} and the fact that $\|U_t^{-1}\|\leq 1$ we have 
    \begin{align}
        \left\| \nabla^{-2} \Phi_{t+1}(\u_t) -  \frac{1-\|\z_t\|^2}{2d\eta}  \sum_{k=1}^{m+1} (-\alpha_t)^{k-1} U_t^{-k} \right\|\leq \frac{(R^2-\|\z_t\|^2)\cdot c^m}{2\nu  \cdot(1-c)}.\nn
    \end{align}
    Now, the fact that $\frac{R^2-\|\z_t\|^2}{2d\eta}  \sum_{k=1}^{m+1} (-\alpha_t)^{k-1} U_t^{-k} =\sum_{k=1}^{m+1} \gamma_t^{k-1} \Sigma_t^{k}$ completes the proof.
\end{proof}
The key to the computational efficiency of our approach lies in the fact that we only need to compute the full inverse of a matrix in a small fraction of the rounds, as demonstrated by the following lemma. The proof of this lemma leverages the stability of the Newton iterates, which is ensured by the non-linear terms in $(\Phi_t)$.

%

\subsection{Number of Taylor Expansion Points (Proof of \cref{lem:movement})}
	\label{sec:movement_proof}
	For the proof of \cref{lem:movement}, we need the following elementary result.
	\begin{lemma}
		\label{lem:damped}
		Let $\nu>0$ and $R>0$ be given and define $\Psi(\x)\coloneqq - \nu \ln (R^2-\|\x\|^2)$. For any $\u, \w\in \cB(R)$, we have
		\begin{align}
			\frac{1}{\nu} \|\w  - \u \|^2_{\nabla^2\Psi(\w )} \geq \frac{(\|\w \|^2-\|\u\|^2)^2}{(R^2-\|\w \|^2)^2}.\nn 
		\end{align}
	\end{lemma}

	\begin{proof}
		Fix $\u,\w\in \cB(R)$. We have
		\begin{align}
			\frac{1}{2\nu}	\|\w -\u\|^2_{\nabla^2 \Psi(\w )} & = (\w-\u)^\top \left(\frac{I}{R^2- \|\w\|^2} + \frac{2 \w\w^\top}{(R^2-\|\w\|^2)^2} \right) (\w-\u) ,\nn \\&=\frac{\|\w \|^2+ \|\u\|^2 - 2 \w ^\top \u  - 2 \|\w \|^2 \w ^\top \u +\|\w \|^4  +2(\w ^\top \u)^2 - \|\w \|^2 \|\u\|^2}{(R^2-\|\w\|^2)^2},\nn \\
			& = \frac{2(\|\w\|^2-\w ^\top\u)(R^2-\w ^\top\u)}{(R^2-\|\w \|^2)^2} +\frac{\|\u\|^2- \|\w \|^2}{R^2-\|\w \|^2},\nn \\
			& = \frac{2(\|\w\|^2-\w ^\top\u)(R^2-\|\w \|^2)}{(R^2-\|\w \|^2)^2} + \frac{2(\|\w\|^2-\w ^\top\u)^2}{(R^2-\|\w \|^2)^2}  +\frac{\|\u\|^2- \|\w \|^2}{R^2-\|\w \|^2}.\nn 
		\end{align}
		Now using that $-\w \u=2^{-1}(\|\w -\u\|^2 - \|\w \|^2 - \|\u\|^2)$, we get that 
		\begin{align}
			\frac{1}{2\nu}	\|\w -\u\|^2_{\nabla^2 \Psi(\w )} 	& = \frac{\|\w- \u\|^2+\|\w \|^2-\|\u\|^2 }{R^2-\|\w \|^2} + \frac{2(\|\w\|^2-\w ^\top\u)^2}{(R^2-\|\w \|^2)^2}  +\frac{\|\u\|^2- \|\w \|^2}{R^2-\|\w \|^2},\nn \\
			& = \frac{\|\w- \u\|^2}{R^2-\|\w \|^2} + \frac{(\|\w- \u\|^2+\|\w \|^2-\|\u\|^2 )^2}{2(R^2-\|\w \|^2)^2}.\label{eq:delta} 
		\end{align}
		Now, consider the function \[f\colon X \rightarrow \frac{X}{R^2-\|\w \|^2} +\frac{(X+\|\w \|^2 -\|\u\|^2)^2}{2(R^2-\|\w \|^2)^2}.\] Note that $\text{sgn}(f'(X))=\text{sgn}(X-\|\u\|^2+1)$. Thus, since $\|\u\|^2\leq 1$, the function $f$ is non-decreasing over $\reals_{\geq 0}$, and so $f(\|\w-\u\|^2)\geq f(0)$. Using this with \eqref{eq:delta}, we get 
		\begin{align}
			\frac{1}{\nu}	\|\w -\u\|^2_{\nabla^2 \Psi(\w )} 	 \geq  \frac{(\|\w \|^2-\|\u\|^2)^2}{(R^2-\|\w \|^2)^2}.\nn
		\end{align}
	\end{proof}

	\begin{proof}[Proof of \cref{lem:movement}]
Let $i_1, \dots, i_n$ be the rounds $t$ where $\z_t \neq \z_{t-1}$, and note that by \cref{line:mark} of \cref{alg:pseudo}, we have 
		\begin{align}
			|\|\z_{i_{k+1}}\|^2 - \|\z_{i_k}\|^2| > c \cdot (R^2-\|\z_{i_k}\|^2), \quad \forall k \in[n-1].	\label{eq:sitone}
		\end{align}
		Further, let 
		\begin{align}
			\alpha_t \coloneqq \frac{\|\u_{t+1}\|^2 -\|\u_{t}\|^2}{R^2-\|\u_{t+1}\|^2}, \quad \text{and} \quad \mu_t \coloneqq \frac{\|\u_{t}\|^2 -\|\u_{t+1}\|^2}{R^2-\|\u_{t}\|^2}.\nn
		\end{align}
		Fix $k\in[n-1]$. Suppose that $(\sum_{t=i_{k}}^{i_{k+1}-1} \alpha_t )\vee (\sum_{t=i_{k}}^{i_{k+1}-1} \mu_t)\leq 1/2$ and let $m_k \coloneqq i_{k+1}-i_k$. In this case, by \eqref{eq:sitone} we have that 
		\begin{align}
			\ln(1+c)	&\leq  \left(\ln \frac{R^2-\|\z_{i_{k}}\|^2}{R^2-\|\z_{i_{k+1}}\|^2} \right) \vee  \left(\ln \frac{R^2-\|\z_{i_{k+1}}\|^2}{R^2-\|\z_{i_{k}}\|^2}\right), \nn \\ &\leq  \left(\ln \prod_{t=i_k}^{i_{k+1}-1} (1+\alpha_t)\right) \vee  \left(\ln \prod_{t=i_k}^{i_{k+1}-1} (1+\mu_t)\right)  ,\nn \\ &= \left(\sum_{t=i_k}^{i_{k+1}-1} \ln (1+\alpha_t)\right) \vee \left(\sum_{t=i_k}^{i_{k+1}-1} \ln (1+\mu_t)\right),\nn \\
			& \leq \ln \left(1+ \frac{1}{m_k}\sum_{t=i_k}^{i_{k+1}-1} \alpha_t  \right)^{m_k}\vee \ln \left(1+ \frac{1}{m_k}\sum_{t=i_k}^{i_{k+1}-1} \mu_t  \right)^{m_k}, \nn \quad \text{(Jensen)} \\
			& \leq \ln \left(1 +2 \sum_{t=i_k}^{i_{k+1}-1} \alpha_t\right)\vee \ln \left(1 +2 \sum_{t=i_k}^{i_{k+1}-1} \mu_t\right), \nn
		\end{align} 
		where the last inequality follows by the facts that $(\sum_{t=i_{k}}^{i_{k+1}-1} \alpha_t )\vee (\sum_{t=i_{k}}^{i_{k+1}-1} \mu_t)\leq 1/2$ and $(1+x)^r \leq 1+\frac{r x}{1-(r-1)x}$, for all $x\in(-1,\frac{1}{r-1}]$ and $r\geq 1$. Now, using that $\ln(1+x)\leq x$ for $x\geq 0$ and $\ln (1+x)\geq x/2$, for $x\in(0,1)$, we get that 
		\begin{align}
			\frac{c}{2}\leq\ln(1+c)& \leq  \left(2 \sum_{t=i_k}^{i_{k+1}-1} \alpha_t\right)\vee \left(2\sum_{t=i_k}^{i_{k+1}-1} \mu_t\right),\label{eq:new} \\
			& \leq  2\left(\sqrt{m_k \sum_{t=i_k}^{i_{k+1}-1} \alpha^2_t}\right)\vee \left(\sqrt{m_k\sum_{t=i_k}^{i_{k+1}-1} \mu^2_t}\right),\nn  \quad (\text{Jensen}) \\
			&  \leq  2 \sqrt{m_k  \sum_{t=i_k}^{i_{k+1}-1} \alpha^2_t +  m_k  \sum_{t=i_k}^{i_{k+1}-1} \mu^2_t}. 
			\label{eq:checkpoint}
		\end{align}
		So far, we have assumed that $(\sum_{t=i_{k}}^{i_{k+1}-1} \alpha_t )\vee (\sum_{t=i_{k}}^{i_{k+1}-1} \mu_t)\leq 1/2$. If this does not hold, then we have $(\sum_{t=i_{k}}^{i_{k+1}-1} \alpha_t )\vee (\sum_{t=i_{k}}^{i_{k+1}-1} \mu_t)\geq 1/2$. This implies \eqref{eq:new} from which \eqref{eq:checkpoint} follows. Now, \eqref{eq:checkpoint} implies 
		\begin{align}
			\sum_{t=i_k}^{i_{k+1}-1} \alpha^2_t +\sum_{t=i_k}^{i_{k+1}-1} \mu^2_t\geq \frac{c^2}{16 m_k}.\nn
		\end{align}
		Thus, by summing over $k=1,\dots,n-1$, and using \cref{lem:damped} and \cref{lem:close} (in particular \eqref{eq:sum}), we get 
		\begin{align}
			\frac{1}{2}+ \frac{13^2d  \ln (1+  T/d)}{\nu \eta } \geq \sum_{t=1}^T ( \alpha^2_t + \mu_t^2 )\geq \sum_{k=1}^n  \frac{c^2}{16 m_k} =\sum_{k=1}^n  \frac{c^2}{16 (i_{k+1}-i_k)}\geq \frac{c^2 n^2}{16 T},\nn 
		\end{align}
		where the last inequality follows by the fact that $x\mapsto 1/x$ is convex and Jensen's inequality.
		By taking the square-root on both sides and rearranging, we get that 
		\begin{align}
			n \leq  \frac{52}{c} \sqrt{ T \cdot\left(1+\frac{d\ln (1+ T/d)}{\nu\eta}\right)}.
		\end{align}
	\end{proof}

\subsection{Structural Results for \barons{} and \ftrl{}}
Most of the structural results we present in this section are small modifications of existing results in \citep{mhammedi2023quasi}.
\label{sec:structural}
\begin{lemma}
    \label{lem:selfconcord}For any $t\geq 1$,
    the functions $\Psi$ and $\Phi_t$ in \eqref{eq:Phi} are self-concordant with constant $1/\sqrt{\nu}$.
\end{lemma}

\begin{proof}Fix $t\geq 1$. First, we note that $\x\mapsto \Psi(\x)/\nu=  -\ln (R^2- \|\x\|^2)$ is self-concordant with constant $1$ (see e.g.\,\cite[Exampled 5.1.1]{nesterov2018lectures}). Thus, $\Psi$ is a self-concordant function with constant $1/\sqrt{\nu}$; this follows by the fact that if a function $f$ is self-concordant with constant $M_f$, then $\alpha f$, for $\alpha >0$, it is self-concordant with constant $1/\sqrt{\alpha}$ (see e.g.\,\cite[Corollary 5.1.3]{nesterov2018lectures}). On the other hand, since $\Phi_t(\x)$ is equal to $\Psi(\x)$ plus a quadratic in $\x$, then $\Phi_t$ is self-concordant with the same constant as $\Psi$ (see e.g.\,\cite[Corollary 5.1.2]{nesterov2018lectures}).
\end{proof}

\begin{lemma}
    \label{lem:gradsum}
    Let $\eta,\nu \in(0,1)$, $R>0$, and $\Gtilde>0$ be such that $\nu \geq 10 \Gtilde R$ and $\eta \leq \frac{1}{5 \Gtilde R}$. Further, let $(\gt_t)\subset \reals^d$ be a sequence of vectors such that $\|\gt_t\|\leq \Gtilde$, for all $t\geq 1$. Then, for any sequence $(\y_t) \subset \inte\, \cB(R)$, the potential functions $(\Phi_t)$ in \eqref{eq:Phi} satisfy 
    \begin{gather}
        \forall t\geq 1, \quad  \|\gt_t\|^{2}_{\nabla^{-2}\Phi_{t}(\y_t)} \leq \frac{\Gtilde^2R^2}{2\nu}, \label{eq:gradientbound}
        \shortintertext{and}
       \forall T \geq 1,\quad  \sum_{t=1}^T \|\gt_t\|^{2}_{\nabla^{-2}\Phi_{t}(\y_t)} \leq \frac{ d \ln (1+T/d)}{\eta}.\label{eq:inee}
    \end{gather}
\end{lemma}
\begin{proof}
	Fix the sequence $(\y_t)$.
    First, note that the Hessian of $\Psi$ in \eqref{eq:Phi} satisfies 
        \begin{align}
      \forall t\geq 1,\quad    \nabla^2 \Psi(\y_t) = \frac{2\nu}{R^2- \|\y_t\|^2} I + \frac{4\nu \y_t \y_t^\top}{(R^2- \|\y_t\|^2)^2} I.  \label{eq:hessian}
        \end{align}
Therefore, we have 
\begin{align}
    \forall t\geq 1,\quad \|\gt_t\|^2_{\nabla^{-2} \Phi_t(\y_t)}&\leq \gt_t^\top \left( \nabla^2 \Psi(\y_t) \right)^{-1}\gt_t, \nn \\
    & \leq \gt_t^\top \left( \nabla^2 \Psi(\y_t) \right)^{-1} \gt_t, \nn \\
    & \leq \frac{R^2 \Gtilde^2}{2\nu},
\end{align}
where in the last step we used \eqref{eq:hessian} and $(\gt_t)\subset \bbB(\Gtilde)$. This shows \eqref{eq:gradientbound}.

		We now show \eqref{eq:inee}. Since $\eta \leq \frac{1}{5 \Gtilde R}$, $\nu \geq 10 \Gtilde R$, and $(\gt_t) \subset \bbB(\Gtilde)$, we have 
        \begin{align}
       \forall t\geq 1,\quad      \eta \gt_t\gt_t^\top \preceq \frac{\Gtilde}{5 R} I \preceq  \frac{\nu}{5 R^2} I \preceq \frac{\nu}{5(R^2- \|\y_t\|^2)} I.\label{eq:ineq}
        \end{align}
        Combining this with \eqref{eq:hessian} implies that 
        \begin{align}
            \forall t\geq 1,\quad      \eta \gt_t\gt_t^\top \preceq \frac{1}{10}\nabla^2 \Psi(\y_t). \label{eq:maybe}
        \end{align}
        Note that \eqref{eq:hessian} also implies that 
        \begin{align}
        \forall t\geq 1, \quad     \frac{\nu}{R^2} I \preceq \frac{1}{2} \nabla^2 \Psi(\y_t). \label{eq:lowerhess}
        \end{align}
        Therefore, we have
        \begin{align}
       \forall t\geq 1, \quad  \|\gt_t\|^{2}_{\nabla^{-2}\Phi_t(\y_t)}& =  \gt_t^\top\left( \nabla^2 \Psi(\y_t) +\eta \sum_{s=1}^{t-1} \gt_s \gt_s^\top \right)^{-1} \gt_t ,\nn \\
        & \leq \gt_t^\top\left( \frac{1}{2}\nabla^2 \Psi(\y_t) +\eta \sum_{s=1}^{t} \gt_s \gt_s^\top \right)^{-1} \gt_t,\quad \text{(by \eqref{eq:maybe})} \\
        & \leq \gt_t^\top\left(\frac{\nu}{R^2} I +\eta \sum_{s=1}^{t} \gt_s \gt_s^\top \right)^{-1} \gt_t,\quad \text{(by \eqref{eq:lowerhess})} \\
        & \leq \frac{1}{\eta} \gt_t^\top Q_{t}^{-1} \gt_t. \label{eq:didit}
		\end{align}
		where $Q_t \coloneqq \frac{\nu}{R^2 \eta}  I  +\sum_{s=1}^t \gt_s\gt_s^\top$. Thus, by \eqref{eq:didit} and \cite[Lemma 11]{hazan2007logarithmic}, we have 
		\begin{align}
	\forall t\in[T],\quad 		\|\gt_t\|^{2}_{\nabla^{-2}\Phi_t(\y_t)}&\leq 	\frac{1}{\eta} \sum_{t=1}^T \gt_t^\top Q_t^{-1} \gt_t, \nn \\
    &  \leq \frac{1}{\eta}\ln \frac{\det Q_T}{\det Q_0},\nn \\
    & = \frac{1}{\eta} \log \det\left(I + Q_0^{-1} \sum_{t=1}^T \gt_t \gt_t^\top\right),\nn \\
    & \leq  \frac{d}{\eta} \log  \frac{\tr\left( Q_0^{-1}\sum_{t=1}^T \gt_t \gt_t^\top\right)}{d}, \quad \text{(Jensen's inequality)}\nn \\
    &  \leq \frac{d\ln (1+ \frac{\eta TR^2 \Gtilde^2}{\nu d})}{\eta},  \quad \text{(using the expression of $Q_0$ and $(\gt_t)\subset \bbB(\Gtilde)$)}\nn \\
    & \leq \frac{d\ln (1+ \frac{T}{d})}{\eta},\nn
		\end{align}
		where in the last step uses that $\nu \geq \Gtilde R$ and $\eta \leq \frac{1}{\Gtilde R}$. This completes the proof.
	\end{proof}

	\begin{lemma}
		\label{lem:minPhi}
		Let $\eta,R,\Gtilde,\nu>0$, and $T \ge 1$ be given, and suppose that $\nu \leq 10 d \Gtilde R T$ and that $\u_t \in \bbB(R)$ and $\gt_t \in \bbB(\Gtilde)$ for all $t\in[T]$. Then, for any $t\in[T]$, the \ftrl{} iterate $\w_t$ in \eqref{eq:FTRL} satisfies:
		\begin{align}
			\frac{\nu R}{R^2-\|\w_t\|^2} \leq 20d \Gtilde \cdot (1+ 2 \eta \Gtilde R)  T.\nn 
		\end{align} 
	\end{lemma}

	\begin{proof}[Proof]
		Fix $t\in[T]$. Since $\Psi(\x)$ is a self-concordant barrier, we have $\w_t \in \inte\,\bbB(R)$. Thus, by the first-order optimality condition involving $\w_t$, we  have 
		\begin{align}
			\frac{2\nu \w_t}{R^2-\|\w_t\|^2}  + \eta \sum_{s=1}^{t-1} \gt_s \gt_s^\top (\w_t-\u_s) +\sum_{s=1}^{t-1} \gt_s=\bm{0}. \nn 
		\end{align}
		Thus, using that $\w_s,\u_s \in \bbB(R)$ and $\gt_s \in \bbB(\Gtilde)$ for all $s\in[t-1]$, we get 
		\begin{align}
			\frac{2 \nu\|\w_t\|}{R^2-\|\w_t\|^2} \leq \Gtilde \cdot (1+2\eta \Gtilde R)\cdot T. \label{eq:wehad}
		\end{align}
		If $\|\w_t\|\leq R/2,$ then we are done since in this case $\frac{1}{R^2-\|\w_t\|^2}\leq \frac{4}{3R^2}\leq \frac{2}{R^2}$, and so 
        \begin{align}
            \frac{\nu R}{R^2-\|\w_t\|^2} \leq \frac{2 \nu}{R} \stackrel{(a)}\leq 20 d \Gtilde \cdot T \leq 20 d \Gtilde \cdot (1+ 2\eta \Gtilde R)\cdot T,
        \end{align}
        where $(a)$ follows from the assumption that $\nu \leq 10d \Gtilde  R T$.
        Now, suppose that $\|\w_t\|> R/2$. Plugging this into \eqref{eq:wehad} directly implies that 
		\begin{align}
			\frac{\nu R}{R^2-\|\w_t\|^2}\leq\Gtilde \cdot (1+2\eta \Gtilde R) T,
		\end{align}
        which completes the proof.
	\end{proof}
	
	\begin{lemma}
		\label{lem:bio}
		Let $\eta,R,\Gtilde,B,\nu>0$, and $T \ge 1$ be given, and suppose that $\nu \leq 10d \Gtilde  R T$ and that $\u_t \in \bbB(R)$ and $\gt_t \in \bbB(\Gtilde)$ for all $t\in[T]$. Further, for $t\in[T]$, let $\w_t$ be the \ftrl{} iterate in \eqref{eq:FTRL}; that is, $\w_t \in \argmin_{\w\in \bbB(R)} \Phi_t(\w)$. Then, for any $t\in[T]$ and $\z \in \inte\,\bbB(R)$ such that $\|\z - \w_t\|^2_{\nabla^2\Psi(\z)}\leq  \nu  B^2$, where $\Psi(\z) \coloneqq -\nu \cdot \log (R^2 - \|\z\|^2)$, we have
		\begin{gather}
			\|\nabla \Phi_{t}(\z)\| \leq C_T, \nn \\
            \intertext{where $C_T \coloneqq  d\Gtilde \cdot (41+ 40 B) \cdot (1+ 2 \eta \Gtilde R) T$. Furthermore, }
		  \nabla^2 \Phi_{t}(\z)   \preceq \left(\eta \Gtilde^2 T + C_T/R   + 2 \nu^{-1} C_T^2  \right) \cdot I.  
		\end{gather}
	\end{lemma}

	\begin{proof}[Proof]
		Fix $t\in[T]$ and $\z \in \inte\,\bbB(R)$ such that $\|\z - \w_t\|^2_{\nabla^2\Psi(\z)}\leq   \nu B^2$. By \cref{lem:damped}, we have 
		\begin{align}
			B^2 & \geq 	\left(\frac{ \|\z\|^2 - \|\w_t\|^2}{R^2-\|\z\|^2}\right)^2 =  \left(\frac{ R^2- \|\w_t\|^2}{R^2-\|\z\|^2} -1\right)^2.\nn 
		\end{align}
		This implies that
		\begin{align}
			\frac{2\nu}{R^2-\|\z\|^2} \leq \frac{2\nu  \cdot(1+B)}{R^2-\|\w_t\|^2}\leq 40d R^{-1}\Gtilde (1+ B) (1+ 2 \eta \Gtilde R)\cdot T,  \label{eq:RGB}
		\end{align}
		where the last inequality follows by \cref{lem:minPhi}. Therefore, by the expression of $\nabla \Phi_t(\z)$ and the triangle inequality, we have
		\begin{align}
			\|\nabla \Phi_t(\z)\|&\leq \left\|\frac{2\nu  \z}{R^2-\|\z\|^2} \right\|+ \left\| \eta \sum_{s=1}^{t-1} \gt_s \gt_s^\top (\z-\u_s) +\sum_{s=1}^{t-1}\gt_s \right\| ,\nn\\  &\leq \frac{2\nu R}{R^2-\|\z\|^2}+ \Gtilde \cdot (1+ 2 \eta\Gtilde R) T,\nn \\
			& \leq C_T, \quad (C_T \text{ as in the lemma statement}) \nn 
		\end{align}
        where in the last inequality we used \eqref{eq:RGB}.
		On the other hand, we have 
		\begin{align}
			\nabla^2\Phi_t(\z) & = \frac{2 \nu I}{R^2-\|\z\|^2} + \frac{4\nu \z \z^\top}{(R^2-\|\z\|^2)^2}  + \eta \sum_{s=1}^{t-1} \gt_s \gt_s^\top, \nn \\
			& \preceq \left(\eta \Gtilde^2T  + C_T/R  + 2 \nu^{-1} C_T^2  \right) \cdot I .\nn 
		\end{align}
		where  the last inequality follows from \eqref{eq:RGB} and the facts that $\z\in \bbB(R)$ and $\gt_s \in \bbB(\Gtilde)$, for all $s \in [t-1]$.
	\end{proof}
	
    \begin{lemma}[Master lemma]
		\label{lem:close}
		Let $\eta, \nu, c \in(0,1)$, $T\in \mathbb{N}$, and $\Gtilde >0$ be given. Further, let $(\u_t)$ be the iterates of \cref{alg:pseudo} with parameters $(T,\eta, \nu, c)$ and suppose that \begin{itemize}
            \item $\gt_t\in \bbB(\Gtilde)$, for all $t\in [T]$;
            \item $10\Gtilde R  \leq \nu \leq  10d\Gtilde R T $; and  
            \item $\eta  \leq \frac{1}{5 \Gtilde R}$.
        \end{itemize}
       Then, we have $(\u_t)\subset  \inte\,\bbB(R)$ and 
		\begin{gather}
			\forall t \in [T], \ \ 	\frac{\sqrt{\nu}}{4}\left(\|\u_t-\w_t  \|_{\nabla^2 \Phi_t(\u_t)}-2\xi\right) \leq \frac{\sqrt{\nu}}{2} \left(\lambda(\u_t, \Phi_{t})-\xi\right)\leq\lambda(\u_{t-1}, \Phi_{t})^2 \leq \frac{2 \Gtilde^2 R^2}{\nu},\nn
		\end{gather}
		where $\xi \coloneqq \sqrt{\frac{\Gtilde R}{30 T}}$ and $\lambda(\cdot,\cdot)$ is the Newton decrement (see \cref{sec:self-concordant}).
		Further, we have\begin{gather}\sum_{t=1}^T \|\u_t - \w_t\|_{\nabla^2 \Phi_{t}(\u_t)}  \leq \frac{3\sqrt{\nu}}{32} +  \frac{6 d \ln (1+T/d)}{\eta\sqrt{\nu}}.
    \shortintertext{and} 
			\sum_{t=1}^T \|\u_t - \u_{t-1}\|_{\nabla^2 \Psi(\u_t)}^2 + \sum_{t=1}^T \|\u_t - \u_{t-1}\|_{\nabla^2 \Psi(\u_{t-1})}^2 \leq \frac{\nu}{2}+ \frac{13^2 d\ln  (1+T/d)}{\eta}. \label{eq:sum}
		\end{gather}
	\end{lemma}

	\begin{proof}[Proof of \cref{lem:close}] 
	Define
        \begin{gather}
		\utilde_{t+1} \coloneqq \u_t - \nabla^{-2}\Phi_{t+1}(\u_t) \nabla \Phi_{t+1}(\u_t), \quad \text{and} \quad 	 \bnablat_t \coloneqq \sum_{k=1}^{m+1}\left(\frac{2\nu}{R^2-\|\z_t\|^2} - \frac{2\nu}{R^2-\|\u_t\|^2}\right)^{k-1} \Sigma_t^{k} \bnabla_t,\nn
        \end{gather}
		and note that $\u_{t+1}=\u_t-\bnablat_t$ from \cref{line:taylor} of \cref{alg:pseudo}. We will show by induction that for all $s \geq 1$, 
		\begin{gather}
			\u_s\in \mathrm{int}\  \bbB(R),
            \shortintertext{and}
			\frac{\sqrt{\nu }}{4}(\|\u_s-\w_s  \|_{\nabla^2 \Phi_s(\u_s)}-2\xi) \leq \frac{\sqrt{\nu }}{2} (\lambda(\u_s, \Phi_{s})-\xi)\leq \lambda(\u_{s-1}, \Phi_{s})^2 \leq \frac{2 \Gtilde^2 R^2}{\nu}, \label{eq:induction} 
		\end{gather}
		where $\xi =\sqrt{\frac{\Gtilde R}{30 T}}$ and $\u_0 =\mathbf{0}$ by convention. The base case follows trivially since $\nabla \Phi_1(\u_0)=\nabla \Phi_1(\u_1)=\bm{0}$ and $\u_1 =\w_1$. Suppose that \eqref{eq:induction} holds for $s=t$. We will show that it holds for $s=t+1$. By the expression of $\Phi_{t+1}$ in \eqref{eq:Phi}, we have $\nabla \Phi_{t+1}(\u_t)=\gt_t + \nabla \Phi_t(\u_t)$, and so by the fact that $(a+b)^2 \leq 2 a^2 +2 b^2$, we get 
		\begin{align}
			\lambda(\u_t, \Phi_{t+1})^2 &= \|\nabla \Phi_{t+1}(\u_t)\|^2_{\nabla^{-2}\Phi_{t+1}(\u_t)}  ,\nn \\ &\leq 2   \|\nabla \Phi_t(\u_t)\|^2_{\nabla^{-2}\Phi_{t}(\u_t)}  + 2 \|\gt_t\|^2_{  \nabla^{-2}\Phi_{t}(\u_t)}, \ \ (\nabla^{2}\Phi_{t+1}(\cdot) \succeq \nabla^2 \Phi_t(\cdot)) \nn \\
			& =2   \lambda(\u_t, \Phi_t)^2  + 2 \|\gt_t\|^2_{  \nabla^{-2}\Phi_{t}(\u_t)}, \label{eq:curz}\\
			& \le 2 \cdot \left(16 \frac{\Gtilde^4 R^4}{\nu^3} + 8 \xi  \frac{\Gtilde^2 R^2}{\nu^{3/2}}  + \xi^2\right)    + \frac{\Gtilde^2 R^2}{\nu},\label{eq:penulpenul}  \\ & \leq  \frac{2 \Gtilde^2 R^2}{\nu},
			\label{eq:postcurz} 
		\end{align}
		where in \eqref{eq:penulpenul} we used the induction hypothesis in \eqref{eq:induction} for $s=t$ and the bound on $\|\gt_t\|^2_{  \nabla^{-2}\Phi_{t}(\u_t)}$ from \cref{lem:gradsum}; and \eqref{eq:postcurz} uses that $10 \Gtilde R\leq \nu\leq 10 d\Gtilde R T$ and $\xi=\sqrt{\frac{\Gtilde R}{30 T}}$ (which implies that $\xi^2 \leq \frac{\Gtilde^2 R^2}{3\nu}$).
        
        By taking the square-root in \eqref{eq:postcurz}, we get that 
        \begin{align}
            \lambda(\u_t, \Phi_{t+1}) \leq \frac{\sqrt{2} \Gtilde R}{\sqrt{\nu}}.\label{eq:blue}
        \end{align}
        Using this with \cref{lem:properties} and the facts that $\utilde_{t+1}$ is the standard Newton step and $\Phi_{t+1}$ is self-concordant with constant $1/\sqrt{\nu}$, we get that \begin{align}
            \lambda(\utilde_{t+1},\Phi_{t+1}) &\leq  \frac{\nu^{-1/2}\cdot \lambda(\u_t,\Phi_{t+1})^2 }{(1- \nu^{-1/2} \lambda (\u_t, \Phi_{t+1}))^2}, \label{eq:preinterim}\\
            & \leq \frac{\nu^{-1/2}\cdot \lambda(\u_t,\Phi_{t+1})^2 }{(1- \sqrt{2}\nu^{-1} \Gtilde R)^2}, \quad \text{(by \eqref{eq:postcurz})}\nn \\
            & \leq \frac{2}{\sqrt{\nu}}\cdot \lambda(\u_t,\Phi_{t+1})^2, \label{eq:interm}
		\end{align} 
        where the last inequality follows by the fact that $\nu \geq 10\Gtilde R$. Using \eqref{eq:interm} together with \eqref{eq:interm} and \eqref{eq:blue} implies 
        \begin{align}
            \lambda(\utilde_{t+1},\Phi_{t+1}) \leq \frac{4\Gtilde^2 R^2}{\nu^{3/2}} \leq \frac{2}{3\sqrt{6}} \sqrt{\Gtilde R} \leq \frac{\sqrt{\nu}}{9},  \label{eq:unsel}
        \end{align}
		where we used the fact that $\nu \geq 10\Gtilde R$ again.
		Using this with \cref{lem:properties} and the facts that $\w_{t+1}$ is the minimizer of $\Phi_{t+1}$ and $\Phi_{t+1}$ is self-concordant with constant $1/\sqrt{\nu}$ (see \cref{lem:selfconcord}), we have \begin{align}\|\utilde_{t+1}- \w_{t+1}\|_{\nabla^2 \Phi_{t+1}(\utilde_{t+1})}\leq 2 \lambda (\utilde_{t+1}, \Phi_{t+1}).
        \end{align} Combining this with \eqref{eq:unsel} implies that \begin{align}\|\utilde_{t+1}- \w_{t+1}\|^2_{\nabla^2 \Phi_{t+1}(\utilde_{t+1})}\leq \frac{4 \nu}{81}
        \end{align} Thus, \cref{lem:bio} instantiated with $B = 2/9$ implies that  
		\begin{gather}
			\label{eq:theprec}
			\nabla^2 \Phi_{t+1}(\utilde_{t+1})   \preceq (\eta \Gtilde^2T + C_T/R + 2 \nu^{-1 }C_T^2 ) \cdot I,
            \shortintertext{where}
            C_T\coloneqq 51d\Gtilde \cdot   (1+ 2 \eta \Gtilde R) T. \label{eq:CT}
		\end{gather}
		On the other hand, since $\bnabla_t = \nabla \Phi_{t+1}(\z_t)$ we have,
		\begin{align}
			\|\utilde_{t+1}- \u_{t+1}\|&= \left\| \nabla^{-2}\Phi_{t+1}(\u_t) \nabla \Phi_{t+1}(\u_t)- \sum_{k=1}^{m+1}\left(\frac{2\nu}{R^2-\|\z_t\|^2} - \frac{2\nu}{R^2-\|\u_t\|^2}\right)^{k-1} \Sigma_t^{k} \bnabla_t \right\|,\nn \\ &=  \left\|\left( \nabla^{-2}\Phi_{t+1}(\u_t)  - \sum_{k=1}^{m+1}\left(\frac{2\nu}{R^2-\|\z_t\|^2} - \frac{2\nu}{R^2-\|\u_t\|^2}\right)^{k-1} \Sigma_t^{k}\right)  \nabla \Phi_{t+1}(\u_t)\right\|,\nn \\& \leq  \left\| \nabla^{-2}\Phi_{t+1}(\u_t)  - \sum_{k=1}^{m+1}\left(\frac{2\nu}{R^2-\|\z_t\|^2} - \frac{2\nu}{R^2-\|\u_t\|^2}\right)^{k-1} \Sigma_t^{k} \right\|\cdot\|\nabla \Phi_{t+1}(\u_t)\| .\label{eq:miso}
		\end{align}
        Now, by the induction hypothesis (i.e., \eqref{eq:induction}), we have $\|\u_t-\w_t  \|_{\nabla^2 \Phi_t(\u_t)}\leq 8\nu^{-1/2}\xi + \frac{8 \Gtilde^2 R^2}{\nu^{3/2}} \leq \frac{2 \sqrt{\nu}}{9}$, where the last inequality uses that $\nu \geq 10\Gtilde R$ and $\xi = \sqrt{\frac{\Gtilde R}{30 T}}$. Thus, by \cref{lem:bio} instantiated with $B = 2/9$, we have \[\|\nabla \Phi_t(\u_t)\|\leq C_T,\] where $C_T$ is as in \eqref{eq:CT}. Therefore, by the triangle inequality and $\gt_t\in \bbB(\Gtilde)$ (by assumption), we have
        \begin{align}
            \|\nabla \Phi_{t+1}(\u_t)\| = \|\nabla \Phi_{t}(\u_t)+ \gt_t\| \leq  C_T + \Gtilde. \label{eq:inter}
        \end{align} 
    On the other hand, by the fact that $\u_t \in \inte\, \bbB(R)$ (by the induction hypothesis), \cref{lem:hessian} implies that 
    \begin{align}
   \left\| \nabla^{-2}\Phi_{t+1}(\u_t)  - \sum_{k=1}^{m+1}\left(\frac{2\nu}{R^2-\|\z_t\|^2} - \frac{2\nu}{R^2-\|\u_t\|^2}\right)^{k-1} \Sigma_t^{k} \right\| \leq \frac{R^2 c^m}{2 \nu \cdot (1-c)}.
    \end{align}
    Plugging this and \eqref{eq:inter} into \eqref{eq:miso} implies that
\begin{align}
    \|\utilde_{t+1}- \u_{t+1}\|    & \leq \frac{R^2 c^m\cdot (C_T+\Gtilde) }{2\nu\cdot(1-c)}. \label{eq:yay}
\end{align}
 Combining \eqref{eq:yay} with \eqref{eq:theprec} implies that 
		\begin{align}
			\|\utilde_{t+1}- \u_{t+1}\|_{\nabla^2\Phi_{t+1}(\utilde_{t+1})} \leq \tilde\xi \coloneqq 	\frac{ R^2 c^m\cdot (C_T+\Gtilde) \cdot \sqrt{\eta \Gtilde^2 T + C_T/R + 2 \nu^{-1 }C_T^2} }{2\nu\cdot(1-c)}. \label{eq:bound}
		\end{align}
		We now show that this implies that $\u_{t+1} \in \bbB(R)$. First, by \eqref{eq:blue} and the fact that $\nu \geq 10\Gtilde R$, we have 
		\begin{align}\|\utilde_{t+1} -\u_t\|_{\nabla^2 \Phi_{t+1}(\u_t)}=\lambda(\u_t, \Phi_{t+1})\leq \frac{\sqrt{2} \Gtilde R}{\sqrt{\nu}}\leq \sqrt{\frac{\Gtilde R}{5}} < \frac{\sqrt{\nu}}{7}.
			\label{eq:tilde}
		\end{align} 
        Combining this with \cref{lem:deakin} and the facts that $\Phi_{t+1}$ is self-concordant with constant $1/\sqrt{\nu}$ (\cref{lem:selfconcord}) and $\u_t\in \inte\,\bbB(R)$ (induction hypothesis), we get that \[\utilde_{t+1}\in \inte\, \bbB(R).\]
        Now, by our choice of $m$ in \cref{alg:pseudo} (i.e., $m=\mathfrak{c}\cdot \log_c(dT)$ with $\mathfrak{c}$ a sufficiently large constant) and the facts that $\nu \geq 10\Gtilde R$ and $\eta \leq \frac{1}{5 \Gtilde R}$, we have \begin{align}\tilde\xi &< \frac{1}{5}\sqrt{\frac{\Gtilde R}{30 T}}= \frac{\xi}{5}, \quad \left(\text{since $\xi = \sqrt{\tfrac{\Gtilde R}{30T}}$}\right)\label{eq:xi} \\
	&\leq \frac{\sqrt{\nu}}{86},\label{eq:epsprime}
        \end{align} and so \eqref{eq:bound} implies that \begin{align}\|\utilde_{t+1}- \u_{t+1}\|_{\nabla^2\Phi_{t+1}(\utilde_{t+1})} < \sqrt{\nu }/4. 
			\label{eq:dani}
		\end{align} %
		We now use this to bound the Newton decrement $\lambda(\u_{t+1},\Phi_{t+1})$. By \cref{lem:deakin}, \eqref{eq:dani}, and the fact that $\Phi_{t+1}$ is self-concordant with constant $1/\sqrt{\nu}$ (\cref{lem:selfconcord}), we have 
		\begin{align}
			\|\utilde_{t+1}- \u_{t+1}\|_{\nabla^2\Phi_{t+1}(\u_{t+1})} \leq 2 \|\utilde_{t+1}- \u_{t+1}\|_{\nabla^2\Phi_{t+1}(\utilde_{t+1})} &\leq 2 \tilde\xi,\quad \text{(by \eqref{eq:bound})} \label{eq:rand} \\ &<\sqrt{\nu }/2.  \quad \text{(by \eqref{eq:epsprime})}\label{eq:hessiancomp}
		\end{align}
		Using this and the triangle inequality, we get 
		\begin{align}
			&\lambda(\u_{t+1},\Phi_{t+1})\nn \\ &= \|\nabla \Phi_{t}(\u_{t+1})\|_{\nabla^{-2}\Phi_{t+1}(\u_{t+1})}\nn \\ &  \leq \|\nabla \Phi_{t}(\utilde_{t+1})\|_{\nabla^{-2}\Phi_{t+1}(\u_{t+1})} + \|\nabla \Phi_{t+1}(\u_{t+1})- \nabla \Phi_{t+1}(\utilde_{t+1})\|_{\nabla^{-2}\Phi_{t+1}(\u_{t+1})},\quad (\text{triangle inequality}) \nn \\
			& \leq \|\nabla \Phi_{t}(\utilde_{t+1})\|_{\nabla^{-2}\Phi_{t+1}(\u_{t+1})}  + 2\|\utilde_{t+1}- \u_{t+1}\|_{\nabla^2\Phi_{t+1}(\u_{t+1})},\quad \text{(by \eqref{eq:hessiancomp} and \cref{lem:inter0})}\nn \\
			& \leq (1-2\tilde\xi/\sqrt{\nu })^{-1} \|\nabla \Phi_{t}(\utilde_{t+1})\|_{\nabla^{-2}\Phi_{t+1}(\utilde_{t+1})} +2\|\utilde_{t+1}- \u_{t+1}\|_{\nabla^2\Phi_{t+1}(\u_{t+1})},\quad \text{(by \eqref{eq:rand} and \cref{lem:hessians})} \nn \\
            & \leq (1-2\tilde\xi/\sqrt{\nu })^{-1} \|\nabla \Phi_{t}(\utilde_{t+1})\|_{\nabla^{-2}\Phi_{t+1}(\utilde_{t+1})}  + 4 \tilde\xi, \quad \text{(by \eqref{eq:rand})} \nn \\
            & \leq \lambda(\utilde_{t+1}, \Phi_{t+1}) + 4\tilde\xi \cdot \lambda(\utilde_{t+1}, \Phi_{t+1})/\sqrt{\nu } + 4 \tilde\xi, \quad \text{(\eqref{eq:epsprime} and $\tfrac{1}{1-x}\leq 1 +2 x, \forall x\leq \tfrac{1}{2}$)} \nn \\
			& \leq \lambda(\utilde_{t+1}, \Phi_{t+1}) +  5 \tilde\xi,  \label{eq:proev}
		\end{align}
		where the last inequality follows by \eqref{eq:unsel}. By combining \eqref{eq:proev} with \eqref{eq:interm} and \eqref{eq:postcurz}, we get that 
        \begin{align}
            \lambda(\u_{t+1},\Phi_{t+1}) &\leq 5 \tilde\xi + \frac{2}{\sqrt{\nu}} \lambda(\u_t,\Phi_{t+1})^2,\label{eq:spiegel} \\
            & \leq 5\tilde\xi +\frac{4\Gtilde^2 R^2}{\nu^{3/2}},\label{eq:distant}
		\end{align}
		and so by \eqref{eq:xi} and the fact that $\nu \geq 10 \Gtilde R$ implies that
		\begin{align}
			\lambda(\u_{t+1},\Phi_{t+1}) \leq  \xi  +\frac{2}{\sqrt{\nu}} \lambda(\u_t,\Phi_{t+1})^2 \quad \text{and}	\quad \lambda(\u_{t+1},\Phi_{t+1}) &\leq \frac{\sqrt{\nu}}{4}. \label{eq:inteeee}
		\end{align} 
		Now, by \cref{lem:properties} and the facts that $\w_{t+1}$ is the minimizer of $\Phi_{t+1}$ and $\lambda(\u_{t+1},\Phi_{t+1})\leq \frac{\sqrt{\nu}}2$ (by \eqref{eq:inteeee}), we have $\|\u_{t+1}- \w_{t+1}\|_{\nabla^2 \Phi_{t+1}(\u_{t})}\leq 2 \lambda (\u_{t+1}, \Phi_{t+1})$. Combining this with the inequality on the left-hand side of \eqref{eq:inteeee}, implies \eqref{eq:induction} for $s=t+1$, which concludes the induction. 
		
	We now use \eqref{eq:induction} together with \eqref{eq:curz} to bound the sums \begin{align}S \coloneqq \sum_{t=1}^T \|\u_t - \w_t\|_{\nabla^2 \Phi_{t}(\u_t)}, \ \ S' \coloneqq \sum_{t=1}^T \|\u_t - \u_{t-1}\|_{\nabla^2 \Psi(\u_t)}^2, \ \ \text{and} \ \   S'' \coloneqq \sum_{t=1}^T \|\u_t - \u_{t-1}\|_{\nabla^2 \Psi(\u_{t-1})}^2. \nn
		\end{align} 
        \fakepar{Bounding $S$} We first bound the sum $\sum_{t=1}^T \lambda (\u_t,\Phi_t)^i$, for $i=1,2$. By \eqref{eq:curz} and the fact that $\lambda(\u_{t+1},\Phi_{t+1}) \leq \frac{2}{\sqrt{\nu}}\lambda(\u_t,\Phi_{t+1})^2 +\xi$ (see \eqref{eq:inteeee}), we have 
		\begin{align}
			\lambda(\u_{t+1}, \Phi_{t+1}) \leq \frac{4}{\sqrt{\nu}}	\lambda(\u_{t}, \Phi_{t})^2 +  \frac{4}{\sqrt{\nu}}	 \|\gt_t\|^{2}_{\nabla^{-2}\Phi_{t}(\u_t)}+\xi. \label{eq:match}
		\end{align}
		Summing \eqref{eq:match}, for $t=1,\dots, T$, rearranging, and using that $\lambda(\u_{T+1}, \Phi_{T+1})\geq 0$, we get 
		\begin{align}
			\sum_{t=2}^T \left(\lambda(\u_t, \Phi_t) - \frac{4}{\sqrt{\nu }}  \lambda(\u_t, \Phi_t)^2\right) \leq   \frac{4}{\sqrt{\nu }} \lambda(\u_1, \Phi_1)^2 + \frac{4}{\sqrt{\nu }} \sum_{t=1}^T \|\gt_t\|^{2}_{\nabla^{-2}\Phi_{t}(\u_t)}+T\xi.\nn 
		\end{align}
		Using \eqref{eq:induction} (the induction hypothesis), we have for all $t\in[T]$:
		\begin{align}
		0\leq \frac{4}{\sqrt{\nu}}\lambda(\u_t,\Phi_t)\leq  \frac{16 \Gtilde^2 R^2}{\nu^2}+\frac{4\xi}{\sqrt{\nu}} \leq \frac{1}{4}, \label{eq:lowerbar}
        \end{align}
        where the last inequality follows by the fact that $\nu\geq 10\Gtilde R$ and $\xi =\sqrt{\frac{\Gtilde R}{30 T}} $. Therefore, we have
		\begin{align}
			\frac{3}{4}\sum_{t=1}^T \lambda (\u_t,\Phi_t) &\leq \lambda(\u_1,\Phi_1) + \frac{4}{ \sqrt{\nu }}\sum_{t=1}^T \|\gt_t\|^{2}_{\nabla^{-2}\Phi_{t}(\u_t)},\nn \\ &  \leq  \frac{\sqrt{\nu}}{16} +  \frac{4}{ \sqrt{\nu }} \sum_{t=1}^T \|\gt_t\|^{2}_{\nabla^{-2}\Phi_{t}(\u_t)},\nn \\ & \leq  \frac{\sqrt{\nu}}{16} +  \frac{4 d \ln (1+T/d)}{\eta\sqrt{\nu}}, \label{eq:thesum}
		\end{align}
		where the last inequality follows by \cref{lem:gradsum} and the range assumption on $\eta$. Now, by \cref{lem:properties}, \eqref{eq:lowerbar}, and the facts that $\w_t$ is the minimizer of $\Phi_t$ and $\Phi_t$ is self-concordant with constant $1/\sqrt{\nu}$, we have: \[\|\u_t-\w_t\|_{\nabla^{2}\Phi_t(\u_t)}\leq 2 \lambda(\u_t,\Phi_t).\]
        Combining this with \eqref{eq:thesum} implies that 
        \begin{align}
            S = \sum_{t=1}^T \|\u_t-\w_t\|_{\nabla^2 \Phi_t(\u_t)} \leq \frac{3\sqrt{\nu}}{32} +  \frac{6 d \ln (1+T/d)}{\eta\sqrt{\nu}}.
        \end{align} 
        \fakepar{Bounding $S'$ and $S''$} We now bound $S'$ and $S''$. By \cref{lem:hessians} and the facts that $\|\u_{t+1}-\utilde_{t+1}\|_{\nabla^2 \Psi(\u_{t+1})} \leq \sqrt{\nu}/2$ (which follows from \eqref{eq:hessiancomp} since $\nabla^2 \Psi(\cdot)\preceq \nabla^2 \Phi_{t+1}(\cdot)$) and $\Psi$ is self-concordant with constant $1/\sqrt{\nu}$ (\cref{lem:selfconcord}), we have 
		\begin{align}
			\|\u_{t+1}-\u_t\|_{\nabla^2\Psi(\u_{t+1})} & \leq  	2\|\u_{t+1}-\u_t\|_{\nabla^2\Psi(\utilde_{t+1})}  , \nn \\ &\leq   	2\|\u_{t+1}-\utilde_{t+1}    \|_{\nabla^2\Psi(\utilde_{t+1})} + 2 \|\utilde_{t+1}-\u_t\|_{\nabla^2 \Psi(\utilde_{t+1})}, \quad (\text{triangle inequality}) \nn \\
			& \leq 2\tilde\xi +  2\|\utilde_{t+1}-\u_t\|_{\nabla^2 \Phi_{t+1}(\utilde_{t+1})} \  \  \text{(by \eqref{eq:hessiancomp} and $\nabla^2 \Phi_{t+1}(\utilde_{t+1})\succeq \nabla^2 \Psi(\utilde_{t+1})$)},\nn \\
			\intertext{and so using \cref{lem:deakin} and the facts that $\|\utilde_{t+1}-\u_t\|_{\nabla^2 \Phi_{t+1}(\u_{t})}\leq \sqrt{\nu}/2$ (by \eqref{eq:tilde}) and $\Phi_{t+1}$ is self-concordant with constant $1/\sqrt{\nu}$, we have}
			\|\u_{t+1}-\u_t\|_{\nabla^2\Psi(\u_{t+1})}& \leq 2\tilde\xi +  4\|\utilde_{t+1}-\u_t\|_{\nabla^2 \Phi_{t+1}(\u_{t})},  \nn \\
			& \leq  2\tilde\xi+4\lambda(\u_t, \Phi_{t+1}), \quad  \text{(by \eqref{eq:tilde})} \label{eq:thenew}\\
			& \leq \frac{3\sqrt{\nu}}{5},
		\end{align}
		where the last inequality follows by \eqref{eq:epsprime} and \eqref{eq:tilde}. Thus, since $\Psi$ is self-concordant with constant $1/\sqrt{\nu}$, \cref{lem:deakin} implies that 
		\begin{align}
			\|\u_{t+1}-\u_t\|_{\nabla^2\Psi(\u_{t})}  \leq  \frac{5}{2}\|\u_{t+1}-\u_t\|_{\nabla^2\Psi(\u_{t+1})}.\label{eq:cero} 
		\end{align}
		From this, it suffices to bound the sum $S' \coloneqq \sum_{t=1}^T \|\u_t - \u_{t-1}\|_{\nabla^2 \Psi(\u_t)}^2$. Using \eqref{eq:thenew} and the fact that $(a+b)^2 \leq 5 a^2 + (5/4)b^2$, for all $a,b\in\reals$, we have
		\begin{align}
			\sum_{t=1}^T \|\u_{t+1} - \u_{t}\|_{\nabla^2 \Psi(\u_{t+1})}^2 &\leq 20T\tilde\xi^2  +20\sum_{t=1}^T \lambda(\u_t,\Phi_{t+1})^2, \nn \\
			& \leq 20T\tilde\xi^2 + 40  \sum_{t=1}^T \lambda(\u_t,\Phi_{t})^2+ 40  \sum_{t=1}^T\|\gt_t\|^2_{\nabla^{-2}\Phi_t(\u_t)} ,\quad \text{(by \eqref{eq:curz})}\nn \\
			& \leq 20T\tilde\xi^2 +   40\sum_{t=1}^T \lambda(\u_t,\Phi_{t})^2+ \frac{40 d \ln (1+T/d)}{\eta}, \quad \text{(by \cref{lem:gradsum})}\nn \\
			& \leq 20T\tilde\xi^2 + \frac{5\sqrt{\nu}}{2}\sum_{t=1}^T \lambda(\u_t,\Phi_{t})+ \frac{40 d \ln (1+T/d)}{\eta}, \quad \text{($ \lambda(\u_t,\Phi_{t})\leq \tfrac{\sqrt{\nu}}{16}$ by \eqref{eq:lowerbar})},  \nn \\
			& \leq 20T \tilde\xi^2  + \frac{\nu}{12} +\frac{16 d \log(1+T/d)}{3 \eta} + \frac{40 d \ln (1+T/d)}{\eta},\quad \text{(by \eqref{eq:thesum})}\nn \\
			& \leq   \frac{\nu}{8} + \frac{46 d \ln (1+T/d)}{\eta},
		\end{align}
		where the last inequality follows by the bound on $\tilde\xi$ in \eqref{eq:epsprime}. Combining this with \eqref{eq:cero} implies \eqref{eq:sum}.
	\end{proof}

	\subsection{Regret of \ftrl{} (Proof of \cref{lem:FTRL})}
	\label{sec:FTRL_proof}
\begin{proof} Fix $\w\in \inte\,\bbB(R)$. For any $t\geq 1$, define $\phi_t(\x) \coloneqq \x^\top \gt_t  + \eta  \inner{\gt_t}{\x- \u_t}^2/2$ and $\phi_0(\x) \coloneqq \Psi(\x)$, and note that $\Phi_t(\x)=\sum_{s=0}^{t-1}\phi_s(\x)$ and $\u_t \in \argmin_{\x\in \bbB(R)} \sum_{s=0}^{t-1}\phi_s(\x)$. By \cite[Lemma 3.1]{cesa2006prediction}, we have 
		\begin{align}
			\sum_{t=0}^T \phi_t(\u_{t+1}) \leq \sum_{t=0}^T \phi_t(\w), 
		\end{align}
		which implies that
			\begin{align}
			\sum_{t=1}^T \inner{\u_{t+1}-\w}{\gt_t} & \leq \phi_0(\w) - \phi_0(\u_1) + \frac{\eta}{2} \sum_{t=1}^T \inner{\u_t -\w}{\gt_t}^2, \nn\\ & \leq \Psi(\w) - \Psi(\bm{0}) + \frac{\eta}{2} \sum_{t=1}^T \inner{\u_t -\w}{\gt_t}^2,\label{eq:dude}
		\end{align}
where the last inequality follows by the fact that $\u_1 =\bm{0}\in \argmin_{\x \in \bbB(R)} - \log (R^2 -\|\x\|^2)$.
		Now, it suffices to bound the sum $\sum_{t=1}^T\inner{\u_t-\u_{t+1}}{\gt_t}$. By Taylor's theorem, the exists $\y_t$ in the segment $[\u_t, \u_{t+1}]$ such that
		\begin{align}
			\Phi_{t+1}(\u_t) - \Phi_{t+1}(\u_{t+1}) &\geq   \nabla \Phi_{t+1}(\u_{t+1})^\top(\u_t - \u_{t+1}) +\frac{1}{2} \|\u_t-\u_{t+1}\|^2_{\nabla^2 \Phi_{t+1}(\y_{t})},\nn  \\
			& \geq  \frac{1}{2}\|\u_t-\u_{t+1}\|^2_{\nabla^2 \Phi_{t+1}(\y_{t})}, \label{eq:suppose}
		\end{align}
		where the last inequality uses the fact that $\u_{t+1}\in \argmin_{\x\in \bbB(R)} \Phi_{t+1}(\x)$ is in the interior of $\bbB(R)$ by self-concordance of $\Phi_{t+1}$.
		On the other hand, using the convexity of $\Phi_{t+1}$ and the fact that $\nabla \Phi_{t+1}(\u_t)=\nabla \phi_{t+1}(\u_t)+ \nabla \Phi_{t}(\u_t)=\nabla \phi_{t+1}(\u_{t})$ (by optimality of $\u_t$), we get that
		\begin{align}
			\Phi_{t+1}(\u_t) - \Phi_{t+1}(\u_{t+1}) & \leq   \inner{\u_t-\u_{t+1}}{\nabla \phi_{t+1}(\u_t)}, \nn \\ & =   \inner{\u_t-\u_{t+1}}{\gt_t}(1 + \eta \inner{\gt_t}{\u_t - \u_{t+1}}),
			\nn \\ & \leq  \norm{\u_t - \u_{t+1}}_{\nabla^2\Phi_{t+1}(\y_{t})} \cdot \norm{\gt_t}_{\nabla^{-2}\Phi_{t+1}(\y_{t})}\cdot (1 + \eta \inner{\gt_t}{\u_t - \u_{t+1}}),\nn \\
			& \leq  \frac{3}{2} \norm{\u_t - \u_{t+1}}_{\nabla^2\Phi_{t+1}(\y_{t})} \cdot \norm{\gt_t}_{\nabla^{-2}\Phi_{t+1}(\y_{t})},
		\end{align}
		where the last inequality follows by the fact that $\eta \leq \frac{1}{5 \Gtilde R}$, $(\gt_t)\subset \bbB(\Gtilde)$, and $(\u_t)\subset \bbB(R)$.
		Combining this and \eqref{eq:suppose}, we get \begin{align}
			\norm{\u_t - \u_{t+1}}_{\nabla^2\Phi_{t+1}(\y_{t})} \leq 3\norm{ \gt_t}_{\nabla^{-2}\Phi_{t+1}(\y_{t})} .\nn 
		\end{align}
		Using this and H\"older's inequality leads to 
		\begin{align}
			\inner{\gt_t}{ \u_t - \u_{t+1}} \leq \norm{ \gt_t}_{\nabla^{-2}\Phi_{t+1}(\y_{t})} \norm{\u_t - \u_{t+1}}_{\nabla^{2}\Phi_{t+1}(\y_{t})} &\leq 3\norm{ \gt_t}_{\nabla^{-2}\Phi_{t+1}(\y_t)}^2.\nn  %
		\end{align}
		Thus, by summing this inequality for $t=1,\dots, T$, we get that
		\begin{align}
			\sum_{t=1}^T \inner{\gt_t}{\u_t-\u_{t+1}}&\leq 	3\sum_{t=1}^T\norm{ \gt_t}_{\nabla^{-2}\Phi_{t+1}(\y_t)}^2
			\leq \frac{3 d \ln (d + T/d)}{\eta},
		\end{align}
		where the last inequality follows by \cref{lem:gradsum} and $\nabla^2 \Phi_{t+1}\succeq \nabla^2 \Phi_t$, for all $t\geq 1$.
		Combining this with \eqref{eq:dude}, we get the desired bound.
	\end{proof}

	\subsection{Regret of \barons{} (Proof of \cref{thm:regretNewton})}
	\label{sec:regretNewton_proof}
	\begin{proof}
		First, the fact that $(\u_t)\subset \inte\, \bbB(R)$ follows from \cref{lem:close}.	
		
		We now show \eqref{eq:actual}. Fix $\w\in \inte\, \bbB(R)$ and let $(\w_t)$ be the \ftrl{} iterates in \eqref{eq:FTRL}. We have
		\begin{align}
			&\sum_{t=1}^T \left(\inner{\u_t -\w}{\gt_t}-\frac{\eta}{2} \inner{\u_t -\w}{\gt_t}^2\right) \nn \\ 
				&= \sum_{t=1}^T \left(\inner{\w_t -\w}{\gt_t}-\frac{\eta}{2} (\inner{\w_t -\w}{\gt_t}+\inner{\u_t -\w_t}{\gt_t})^2\right) +\sum_{t=1}^T \inner{\u_t -\w_t}{\gt_t}, \nn \\ 
					&= \sum_{t=1}^T \left(\inner{\w_t -\w}{\gt_t}-\frac{\eta}{2} \inner{\w_t -\w}{\gt_t}^2\right) +\sum_{t=1}^T(1- {\eta}\inner{\w_t -\w}{\gt_t}) \cdot \inner{\u_t -\w_t}{\gt_t}- \frac{\eta}{2}\sum_{t=1}^T \inner{\u_t-\w_t}{\gt_t}^2, \nn \\ 
					&\leq  \sum_{t=1}^T \left(\inner{\w_t -\w}{\gt_t}-\frac{\eta}{2} \inner{\w_t -\w}{\gt_t}^2\right) +\sum_{t=1}^T(1- {\eta}\inner{\w_t -\w}{\gt_t}) \cdot \inner{\u_t -\w_t}{\gt_t}, \nn \\ 
			 &\leq 	\sum_{t=1}^T \left(\inner{\w_t -\w}{\gt_t}-\frac{\eta}{2} \inner{\w_t -\w}{\gt_t}^2\right)  + \frac{7}{5}\sum_{t=1}^T  \|\u_t -\w_t\|_{\nabla^2\Phi_t(\u_t)}  \|\gt_t\|_{\nabla^{-2}\Phi_t(\u_t)}, \label{eq:surrogateregret}
		\end{align}
		where the last step follows by H\"older's inequality, and the facts that $(\gt_t)\subset \bbB(\Gtilde)$, $(\w_t)\subset \bbB(R)$, and $\eta \leq \frac{1}{5 \Gtilde R}$ (which implies that $\eta|\inner{\w_t-\w}{\gt_t}|\leq 2/5$, for all $t\in[T]$).
		
		Now, by \cref{lem:gradsum}, we have $\|\gt_t\|_{\nabla^2 \Phi_t(\u_t)}\leq \frac{\Gtilde R}{\sqrt{2\nu}}$ for all $t\in[T]$, and by \cref{lem:close}, we have 
		\begin{align}
			\sum_{t=1}^T \|\u_t - \w_t\|_{\nabla^2 \Phi_{t}(\u_t)}  \leq \frac{3\sqrt{\nu}}{32} +  \frac{6 d \ln (1+T/d)}{\eta\sqrt{\nu}}.
		\end{align}
		Therefore, we have
		\begin{align}
			\label{eq:firstplug}
			\sum_{t=1}^T  \|\u_t -\w_t\|_{\nabla^2\Phi_t(\u_t)}  \|\gt_t\|_{\nabla^{-2}\Phi_t(\u_t)} & \leq \frac{3\Gtilde R}{32\sqrt{2}} +  \frac{6\Gtilde R d \ln (1+T/d)}{\sqrt{2}\eta \nu}, \nn \\
			& \leq \frac{3\Gtilde R}{32\sqrt{2}} +  \frac{3 d \ln (1+T/d)}{7\eta}, \label{eq:firstterm}
		\end{align}
		where the last inequality follows by $\nu \geq  10 \Gtilde R$. On the other hand, by \cref{lem:FTRL}, we have 
		\begin{align}
			\sum_{t=1}^T \left(\inner{\w_t -\w}{\gt_t}-\frac{\eta}{2} \inner{\w_t -\w}{\gt_t}^2\right) &\leq \Psi(\w) - \Psi(\bm{0})+ \frac{3 d \log (1+T/d)}{\eta}, \nn \\
			& = - \nu \log (1 - \tfrac{\|\w\|^2}{R^2})+ \frac{3 d \log (1+T/d)}{\eta}.
			\label{eq:secondterm}
		\end{align}
		Plugging \eqref{eq:firstterm} and \eqref{eq:secondterm} into \eqref{eq:surrogateregret}, we get 
		\begin{align}
& 	\sum_{t=1}^T \left(\inner{\u_t -\w}{\gt_t}-\frac{\eta}{2} \inner{\u_t -\w}{\gt_t}^2\right)  \leq \frac{21 \Gtilde R}{160\sqrt{2}}- \nu \log (1 - \tfrac{\|\w\|^2}{R^2})+ \frac{18 d \log (1+T/d)}{5 \eta}. \label{eq:dec}
		\end{align}
		Combining this with $\frac{21}{160\sqrt{2}}\leq 1$ implies \eqref{eq:actual}.
				\fakepar{Computational cost} 
				Now, we analyze the computational complexity of \cref{alg:fastexpconcave}, which is equivalent to \cref{alg:pseudo} in that both algorithms produce the same outputs. The most computationally expensive step in \cref{alg:fastexpconcave} occurs in \cref{line:fbi}, where a full matrix inverse is required when $\z_{t} \neq \z_{t-1}$. However, by \cref{lem:movement}, the matrix inverse only needs to be computed at most $O(c^{-1} \sqrt{\frac{d}{\nu\eta} T \ln (1+ T/d)})$ times over $T$ rounds. In the rounds where $\z_{t} = \z_{t-1}$, \cref{alg:fastexpconcave} performs at most $O(m)$ matrix-vector multiplications (see \cref{line:set}-\cref{line:update}), which results in a cost of $O(m d^2)$. Therefore, the total computational complexity of \cref{alg:fastexpconcave} follows from the fact that $m = \mathfrak{c} \cdot \log_c(dT)$, where $\mathfrak{c}$ is a universal constant.
	\end{proof}

\section{OCO Analysis: Proof of \cref{cor:main}}
\label{sec:ocoanalysis}
This appendix provides the proof of \cref{cor:main}. We begin in \cref{sec:instproof} by proving the main reduction result in \cref{lem:inst}, which bounds the instantaneous regret of \cref{alg:projectionfreewrapper} by that of its subroutine $\cA$. In \cref{sec:surro}, we prove the regret bound for \cref{alg:projectionfreewrapper} when $\cA$ is set as the \barons{}, prior to parameter tuning (\cref{lem:meta}). Finally, in \cref{sec:proofoco}, we present the complete proof of \cref{cor:main}.

\subsection{OCO Reduction with Gauge Projections (Proof of \cref{lem:inst})}
\label{sec:instproof}
\begin{proof}
	Fix $t\in [T]$, and let $S_t$, $\bs_t$, $ \u_t,\gt_t$, and ${\w}_t$ be as in \cref{alg:projectionfreewrapper}. We first show that $\w_t \in \K$. By definition of $\w_t$, we have $\w_t = \frac{\u_t}{1+S_t}$. Therefore, by the homogeneity of the Gauge function (see \cref{lem:properties2}), we have 
	\begin{align}
		\gamma_\K(\w_t) &= \frac{\gamma_\K(\u_t)}{1+S_t},\nn \\ &\leq \frac{1 + S_\K(\u_t)}{1+ S_t}, \quad \text{(since $S_\K(\u_t)=\max(0,\gamma_\K(\u_t)-1)$ by \cref{lem:gauge})}\nn \\
		& \leq 1, \label{eq:lessone}
	\end{align}
	where the last inequality follows from $S_\K(\u_t)\leq S_t$ by \cref{lem:approxgrad}. \cref{eq:lessone} implies that $\w_t\in \K$ by definition of the Gauge function (see \cref{def:gaugefunction}).
	
	We now prove the inequality in \eqref{eq:monotone_pre}. For this, define the surrogate loss function $\ell_t$:
	\begin{align}
\forall \w\in \reals^d, \quad 	\ell_t(\w) \coloneqq \inner{\g_t}{\w} -  \mathbb{I}\{\inner{\g_t}{\u_t} <0\} \cdot   \inner{\g_t}{\w_t} \cdot  S_\K(\w).
	\end{align} 
	 Since the pair $(S_t,\bs_t)$ is the output of $\gauged(\K,\u_t,\veps,r)$ with $\veps=1/T$, we have by \cref{lem:approxgrad}:
	\begin{align}
	\forall \u \in \reals^d,\quad S_\K(\u) \geq S_\K(\u_t) + (\u-\u_t)^\top \bs_t - \frac{1}{T}.	\label{eq:subgrad}
	\end{align}
	Now, since $\w_t = \u_t/(1+S_t)$ (see \cref{alg:projectionfreewrapper}) and $S_t \geq S_\K(\u_t)\geq 0$ (by \cref{lem:approxgrad}), we have that $-\mathbb{I}\{\inner{\g_t}{\u_t} <0\} \cdot   \inner{\g_t}{\w_t} \geq 0$. And so, using \eqref{eq:subgrad} and the definition of $\ell_t$, we get 
	\begin{align}
		\forall \u\in \reals^d,\quad \ell_t(\u_t) - \ell_t(\u) &\leq \inner{\g_t - \mathbb{I}\{\inner{\g_t}{\u_t} <0\} \cdot   \inner{\g_t}{\w_t} \cdot \bs_t}{\u_t-\u} + |\inner{\g_t}{\w_t}|\cdot \frac{1}{T} ,\nn \\
		&= \inner{\gt_t}{\u_t -\u} + |\inner{\g_t}{\w_t}|\cdot \frac{1}{T}, \quad \text{(by definition of $\gt_t$ in \cref{alg:projectionfreewrapper})}\nn \\
		& \leq  \inner{\gt_t}{\u_t -\u} +  \frac{GR}{T}, \label{eq:smallone}
	\end{align}
	where the last inequality uses that $\w_t\in \K \subseteq \bbB(R)$ and $\|\g_t\|\leq G$.

			It remains to show that $\inner{\g_t}{\w_t -\u}\leq \ell_t(\u_t) - \ell_t(\u)$, for all $\u \in \K$. First, note that for all $\u \in \K$, we have $S_{\K}(\u)=\max(0,\gamma_\K(\u)-1)=0$ (by \cref{lem:gauge} and the definition of the Gauge function), and so
			\begin{align}
				\ell_t(\u) = \inner{\g_t}\u, \quad \forall \u \in \K. \label{eq:firstequalityperfect}
			\end{align}
			We will now compare $\inner{\g_t}{\w_t}$ to $\ell_t(\u_t)$ by considering cases. Suppose that $S_t =0$. In this case, we have $\w_t = \u_t$ and so $
				\inner{\g_t}{\w_t} = \inner{\g_t}{\u_t} = \ell_t(\u_t).$ Now suppose that $S_t > 0$ and $\inner{\g_t}{\u_t}\geq 0$. In this case, since $\w_t=\frac{\u_t}{1+S_t}$, we immediately have 
			\begin{align}
				\inner{\g_t}{\w_t} \leq \inner{\g_t}{\u_t} = \ell_t(\u_t). \quad [\text{case where } \inner{\g_t}{\u_t}\geq 0] \label{eq:firstsecondperfect}
			\end{align}
			Now suppose that $S_t> 0$ and $\inner{\g_t}{\u_t}<0$. Again, using that $\w_t=\frac{\u_t}{1+S_t}$, we have %
			\begin{align}
			\inner{\g_t}{\w_t}+	\inner{\g_t}{\w_t}\cdot S_{\K}(\u_t) & =   \inner{\g_t}{\u_t} \cdot \frac{1+ S_\K(\u_t)}{1+S_t},\nn \\
			&\leq \inner{\g_t}{\u_t} \cdot \frac{1 + S_t - \frac{1}{T}}{1+ S_t},\quad \text{(since $\inner{\g_t}{\u_t}<0$ and $S_\K(\u_t)\geq S_t -\tfrac{1}{T}$ by \cref{lem:approxgrad})}\nn \\ 
			&\leq  \inner{\g_t}{\u_t} +|\inner{\g_t}{\u_t}| \cdot \frac{1}{T}, \nn \\
			& \leq \inner{\g_t}{\u_t} +\frac{GR}{T},
			\end{align}
			where the last inequality follows from the fact that $\u_t \in \bbB(R)$ and $\|\g_t\|\leq G$.
		Rearranging this, we get
			\begin{gather}	
				\inner{\g_t}{\w_t}  -  \frac{GR}{T} \leq  \inner{\g_t}{\u_t} -  \inner{\g_t}{\w_t} \cdot S_{\K}(\u_t) = \ell_t(\u_t). \quad [\text{case where } \inner{\g_t}{\u_t}< 0] \label{eq:secondperfect}
			\end{gather}
	By combining \eqref{eq:smallone}, \eqref{eq:firstequalityperfect}, \eqref{eq:firstsecondperfect}, and \eqref{eq:secondperfect}, we obtain 
	\begin{align}
	\forall \u \in \K,\quad 	\inner{\g_t}{\w_t -\u} - \frac{GR}{T}  \leq \ell_t(\u_t) - \ell_t(\u)  \leq \inner{\gt_t}{\u_t -\u} + \frac{GR}{T},
	\end{align} which shows the inequality in \eqref{eq:monotone_pre}. 
	
	\fakepar{Bounding the surrogate subgradients} It remains to bound $\|\gt_t\|$ in terms of $\|\g_t\|$. Using that $\gt_t=\g_t - \mathbb{I}\{\inner{\g_t}{\u_t}<0\}\cdot \inner{\g_t}{\w_t}\cdot  \bs_t$ and $\w_t =\frac{\u_t}{1+S_t}$, we have 
			\begin{align}
	\|\gt_t\|& = \|\g_t- \mathbb{I}\{\inner{\g_t}{\u_t} <0\} \cdot   \inner{\g_t}{\w_t} \cdot \bs_t\|,\nn \\&  \leq  	\|\g_t\| + \|\g_t\|\cdot  \frac{\|\u_t\|}{1 + S_t} \cdot  \|\bs_t\|,\quad \text{(by the triangle inequality and Cauchy Schwarz)} \nn \\
	&  \leq	\|\g_t\| \cdot \left(1+  \frac{\| \u_t\|}{r} \right), \quad \text{(since $S_t\geq 0$ and $\|\bs_t\|\leq 1/r$ by \cref{lem:approxgrad})}\nn \\
	&  \leq  (1 +\kappa )\cdot \|\g_t\|, \nn
			\end{align}
			where the last step follows by the assumption that $\u_t\in \bbB(R)$. This completes the proof.
\end{proof}

\subsection{OCO Regret Bound Pre-Tuning of Parameters (Proof of \cref{lem:meta})}
\label{sec:surro}
\begin{proof}
	Fix $\w\in \inte\, \K$. By \cref{lem:inst}, the sequence of loss vectors $(\gt_t)$ that the \barons{} subroutine receives satisfies $(\gt_t)\subset \bbB(\Gtilde)$ with $\Gtilde = 2 \kappa G$. Thus, by invoking the guarantee of \barons{} in \cref{thm:regretNewton}, we get $(\u_t)\subset \bbB(R)$ and
	\begin{align}
	\sum_{t=1}^T \left(\inner{\u_t -\w}{\gt_t}- \frac{\eta}{2} \inner{\u_t -\w}{\gt_t}^2\right)\leq  \Gtilde R -\nu \log (1 - \tfrac{\|\w\|^2}{R^2})+ \frac{4 d \log (1+\frac{T}{d})}{\eta}, \label{eq:onsbound}
	\end{align}
	where we used that $18/5 \leq 4$. We now prove that 
	\begin{align}
		\sum_{t=1}^T \left(\inner{\g_t}{\w_t -\w} -\frac{\eta}{2} \inner{\g_t}{\w_t -\w}^2\right) \leq \sum_{t=1}^T \left(\inner{\gt_t}{\u_t -\w} -\frac{\eta}{2} \inner{\gt_t}{\u_t -\w}^2\right) + 3 GR, \label{eq:boundregret}
	\end{align}
	which together with \eqref{eq:onsbound} would complete the proof. 
	Using that $(\u_t)\subset \bbB(R)$, $(\gt_t)\subset \bbB(2 \kappa G)$, and \cref{ass:setassumption}, we obtain
	\begin{align}
	\forall t\in [T],\forall \u \in \K, \quad |\inner{\gt_t}{\u_t - \u}| \leq 4 \kappa R G.\label{eq:ineqee}
	\end{align} 
	Combining this with the facts that:  
	\begin{itemize}
		\item $\inner{\g_t}{\w_t - \w} \leq \inner{\gt_t}{\u_t  -\w} + \frac{2 GT}{T}$, for all $t\in[T]$ (by \cref{lem:inst});
		\item $x \rightarrow x -\frac{\eta}{2} x^2$ in non-decreasing for all $x \leq \frac{1}{\eta}$ (we instantiate this with $x=\inner{\g_t}{\w_t-\w}$ and $x= \inner{\gt_t}{\u_t - \w}+ \frac{2GR}{T}$); and
	\item $\eta \leq \frac{1}{10 \kappa GR}$;
	\end{itemize}
	we get that for all $t\in[T]$
	\begin{align}
	\inner{\g_t}{\w_t -\w} -\frac{\eta}{2} \inner{\g_t}{\w_t -\w}^2 &\leq \inner{\gt_t}{\u_t -\w} +  \frac{2 G R}{T} -\frac{\eta}{2} \inner{\gt_t}{\u_t -\w}^2 - \frac{2\eta GT}{T} \inner{\gt_t}{\bu_t-\w} -\frac{2\eta G^2R^2}{T^2},\nn \\
		& \leq \inner{\gt_t}{\u_t- \w}-\frac{\eta}{2} \inner{\gt_t}{\u_t- \w}^2+ \frac{3 G R}{T},
	\end{align}
	where the last step follows by \eqref{eq:ineqee} and $\eta\leq \frac{1}{10 \kappa G R}$. Summing this over $t=1,\dots,T$, we obtain \eqref{eq:boundregret}. Combining \eqref{eq:boundregret} with \eqref{eq:onsbound} we get the desired result. 
	\end{proof}

\subsection{Main OCO Regret Bound (Proof of \cref{cor:main})}
\label{sec:proofoco}
\begin{proof}
Fix $\w\in \K$ and let $\wtilde\w \coloneqq \w \cdot (1 - 1/T)$. Note that $\wtilde\w \in \inte\,\K$. By \cref{ass:online} (boundness of $(\g_t)$) and \cref{ass:setassumption} (boundness of $\K$), we have that 
	\begin{align}
		\sum_{t=1}^T \inner{\g_t}{\w_t -\w} &\leq -\sum_{t=1}^T \frac{1}{T}\inner{\g_t}{\w} + \sum_{t=1}^T \inner{\g_t}{\w_t -\wtilde\w}, \nn \\
		& \leq G R + \sum_{t=1}^T \inner{\g_t}{\w_t -\wtilde\w}. \label{eq:wtildegregret0}
	\end{align}
Thus, it suffices bound the (linearized) regret relative to $\wtilde\w$. By instantiating the bound in \cref{lem:meta} with comparator $\wtilde\w \in \inte\,\K$, we get:
\begin{align}
	\sum_{t=1}^T \inner{\g_t}{\w_t -\wtilde\w}  &\leq  \frac{\eta}{2} \sum_{t=1}^T\inner{\g_t}{\w_t -\wtilde\w}^2 +  5\kappa G R -\nu \log (1 - \tfrac{\|\wtilde\w\|^2}{R^2})+ \frac{4 d \log (1+T/d)}{\eta}, \nn \\
	&\leq  \frac{\eta}{2} \sum_{t=1}^T\inner{\g_t}{\w_t -\wtilde\w}^2 +  5\kappa G R +\nu \log T+ \frac{4 d \log (1+T/d)}{\eta},\label{eq:pretogether}
\end{align}
where in the last inequality, we used that \[\|\wtilde\w\| \leq \|\w\| \cdot \left(1-\tfrac{1}{T}\right) \leq R \cdot \left(1-\tfrac{1}{T}\right)\quad \text{and} \quad -\log\left(1 - \left(1-\tfrac{1}{T}\right)^2\right)= - \log \left( \tfrac{2}{T} - \tfrac{1}{T^2}\right) \leq \log T.\]
Now, using that $(\g_t)\subset \bbB(G)$, $(\w_t)\subset \bbB(R)$, and \cref{ass:setassumption} ($\K\subseteq \bbB(R)$), we have $|\inner{\g_t}{\w_t - \wtilde\w}|\leq 2 G R$, for all $t\in[T]$. Combining this with \eqref{eq:pretogether}, we get 
\begin{align}
	\sum_{t=1}^T \inner{\g_t}{\w_t -\wtilde\w}  &\leq  2\eta  G^2 R^2 T +  5\kappa G R +\nu \log T+ \frac{4 d \log (1+\frac{T}{d})}{\eta}. \label{eq:together}
\end{align} 
We now use \eqref{eq:together} to show the desired result. First, note that the optimal tuning of $\eta$ in \eqref{eq:together} is given by 
\begin{align}
\eta^\star = \sqrt{\frac{2d \log (1 + \frac{T}{d})}{T G^2 R^2}}.
\end{align}
We now consider cases.
\fakepar{Case where $\eta^\star\leq \frac{1}{10 \kappa G R}$} First, note that this implies that $\eta = \eta^\star$. Now, using that $\eta^\star\leq \frac{1}{10 \kappa G R}$ and the expression of $\eta^\star$, we have that 
\begin{align}
10\sqrt{2} \cdot \kappa \leq \sqrt{\frac{T}{d\log (1+ \tfrac{T}{d})}}. 
\end{align} 
	This implies that $\nu\leq \sqrt{\frac{2d T}{d\log (1+\frac{T}{d})}}$ (see definition of $\nu$ in \eqref{eq:parameterchoice}). Using this together with $\eta = \eta^\star$ and \eqref{eq:together} implies
	\begin{align}
	\text{(case $\eta^\star\leq \tfrac{1}{10 \kappa G R}$)}\quad	\sum_{t=1}^T \inner{\g_t}{\w_t -\wtilde\w} & \leq  4GR \sqrt{2d T \log \left(1 + \tfrac{T}{d} \right)} + 5 \kappa G R+ GR \sqrt{\frac{2 d T}{ \log(1+ \tfrac{T}{d})}}\cdot  \log T, \nn \\
	 & \leq 5GR \sqrt{2d T \log \left(1 + \tfrac{T}{d} \right)}+ 5 \kappa G R. \label{eq:bounding1}
	\end{align}
	\fakepar{Case where $\eta^\star\geq \frac{1}{10 \kappa G R}$} In this case, we have $\eta = \frac{1}{10 \kappa GR}$. Now, using that $\eta^\star\geq \frac{1}{10 \kappa G R}$ and the expression of $\eta^\star$, we have that 
	\begin{align}
		10\sqrt{2} \cdot \kappa \geq \sqrt{\frac{T}{d\log (1+ \tfrac{T}{d})}}.  \label{eq:cheaper}
	\end{align} 
	This implies that $\nu\leq 20 G R  \kappa d$ (see definition of $\nu$ in \eqref{eq:parameterchoice}). Plugging this and $\eta=\frac{1}{10 \kappa G R}$ into \eqref{eq:together}, we get 
	\begin{align}
		\text{(case $\eta^\star\geq \tfrac{1}{10 \kappa G R}$)}\quad	\sum_{t=1}^T \inner{\g_t}{\w_t -\wtilde\w}  &\leq  \frac{T G R}{5\kappa}  + 5 \kappa G R + 20 G R \kappa d \log T  + 40 d\kappa G R \log(1 + \tfrac{T}{d}),\nn \\
		&\leq  \frac{T G R}{5\kappa}   + 65 d\kappa G R \log(1 + \tfrac{T}{d}),\nn \\
		&\leq  2 GR \sqrt{2 d T \log (1+ \tfrac{T}{d})}    + 65  G R \kappa d\log(1 + \tfrac{T}{d}),\label{eq:bounding2}
	\end{align}
	where the last inequality follows by \eqref{eq:cheaper}. Thus, combining \eqref{eq:bounding1} and \eqref{eq:bounding2}, we get 
	\begin{align}
		\sum_{t=1}^T \inner{\g_t}{\w_t -\wtilde\w} \leq 5 GR \sqrt{2 d T \log (1+ \tfrac{T}{d})}  + 65  G R \kappa d \log(1 + \tfrac{T}{d}).
	\end{align}
	Using this together with \eqref{eq:wtildegregret0} implies the desired result.
\paragraph{Computational cost} By \cref{thm:regretNewton} and the choice of $c=1/2$, the computational cost of the \barons{} subroutine with \cref{alg:projectionfreewrapper} is bounded by $\wtilde{O}\left( d^2 T +  d^{\omega} \sqrt{\tfrac{d T}{\nu \eta}} \right)$.
Now, by the choice of $\eta$ and $\nu$ in \eqref{eq:parameterchoice}, we have $\eta \nu\geq d$. This implies that the computation of the \barons{} subroutine is bounded by 
\[
\wtilde{O}\left(d^2 T +  d^{\omega}\sqrt{T}\right).
\]
Now, in addition to the computational cost of the \barons{} subroutine, \cref{alg:projectionfreewrapper} incurs $O(d) + C_{\gauged}(\cK)$ per round, where $C_{\gauged}(\cK)$ is the cost of one call to the $\gauged$ subroutine (\cref{alg:gauge}) for approximating the gauge distance $S_\K$ and its the subgradients. By \cref{lem:approxgrad}, we have 
\begin{align}
	C_{\gauged}(\cK) \leq \wtilde{O} (1)\cdot  C_{\sep}(\cK).
\end{align}
\end{proof} %
\section{Stochastic Convex Optimization ({Proof of \cref{prop:stoch}})}
\label{sec:stoch}
\begin{proof}
Let $\w^\star \in \argmin_{\u\in\K} f(\u)$. Further, let $\wtilde\w^\star \coloneqq \w^\star \cdot (1 - T^{-1})$. Note that $\wtilde\w^\star \in \inte\,\K$. By \cref{assum:stoch} (boundness of $(\g_t)$), we have that 
\begin{align}
		\sum_{t=1}^T \inner{\g_t}{\w_t -\w^\star} &\leq -\sum_{t=1}^T \frac{1}{T}\inner{\g_t}{\w^\star} + \sum_{t=1}^T \inner{\g_t}{\w_t -\wtilde\w^\star}, \nn \\
		& \leq G R + \sum_{t=1}^T \inner{\g_t}{\w_t -\wtilde\w^\star}. \label{eq:wtildegregret}
	\end{align}
    Now, using Jensen's inequality, we get
  \begin{align}
   \E[f(\widehat{\w}_T)] -f(\w^\star)& \leq \frac{1}{T} \E\left[\sum_{t=1}^T f(\w_t)-f(\w^\star)\right], \nn \\ & \leq \frac{1}{T} \E\left[\sum_{t=1}^T \inner{\gb_t}{\w_t-\w^\star}\right], \quad (\text{by convexity and }\gb_t \in \partial f(\w_t)) \nn \\
   & = \frac{1}{T} \E\left[\sum_{t=1}^T \inner{\g_t}{\w_t-\w^\star}\right]. \quad (\E[\bxi_t]=\bm{0} \text{ by \cref{assum:stoch}}) \label{eq:thats}
   \end{align} 
    Now, by instantiating the bound in \cref{lem:meta} with comparator $\wtilde\w \in \inte\,\K$ and parameters $(\eta, \nu,c)$ as in \eqref{eq:params2}, we get:
\begin{align}
	\sum_{t=1}^T \inner{\g_t}{\w_t -\wtilde\w^\star}  &\leq  \frac{\eta}{2} \sum_{t=1}^T\inner{\g_t}{\w_t -\wtilde\w^\star}^2 +  5 \kappa G R + \nu \log T+  \frac{d \log (1+\frac{T}{d})}{\eta},\label{eq:pretogether2}
\end{align}
where in the last inequality, we used that \[\|\wtilde\w^\star\| \leq \|\w\| \cdot \left(1-\tfrac{1}{T}\right) \leq R \cdot \left(1-\tfrac{1}{T}\right)\quad \text{and} \quad -\log\left(1 - \left(1-\tfrac{1}{T}\right)^2\right)= - \log \left( \tfrac{2}{T} - \tfrac{1}{T^2}\right) \leq \log T.\]
Now, by \cref{assum:stoch} (in particular, the fact that $\g_t = \gb_t + \bxi_t$) together with the fact that $(a+b)^2\leq 2a^2 +2 b^2$ and $\w_t,\wtilde\w^\star \in \bbB(R)$, we have
\begin{align}
    \inner{\g_t}{\w_t- \wtilde\w^\star}^2 \leq 2 \inner{\gb_t}{\w_t - \wtilde\w^\star}^2 + 8 R^2 \|\bxi_t\|^2.  \label{eq:thisonee}
\end{align}
On the other hand, by definition of $\wtilde\w^\star$ 
and the facts that $\w,\w_1,\w_2,\dots \in \bbB(R)$, we have for all $t\in[T]$:
\begin{align}
 2GR   \geq  \inner{\gb_t}{\w_t - \wtilde\w^\star} &\geq  \inner{\gb_t}{\w_t - \w^\star} - \frac{G R}{T}, \nn \\
 & \geq f(\w_t)-f(\w^\star)-\frac{GR}{T},\quad(\text{by convexity of $f$ and $\gb_t\in \partial f(\w_t)$}) \nn \\
 & \geq  - \frac{GR}{T}, \label{eq:lower}
\end{align}
where the last inequality follows by the fact that $\w_t\in \K$ and that $\w^\star$ is the minimizer of $f$ within $\K$. Note that \eqref{eq:lower} implies that for all $t\in[T]$,
\begin{align}
   |\inner{\gb_t}{\w_t - \w^\star}| \leq  \inner{\gb_t}{\w_t - \w^\star} + \frac{2GR}{T}. \label{eq:lower2}
\end{align} 
Picking up from \eqref{eq:thisonee}, we get  
\begin{align}
\sum_{t=1}^T \inner{\g_t}{\w_t-\w^\star}^2 &\leq 2\sum_{t=1}^T \inner{\gb_t}{\w_t - \w^\star}^2 + 8 R^2\sum_{t=1}^T \|\bxi_t\|^2, \nn \\
&\leq  4 GR \sum_{t=1}^T |\inner{\gb_t}{\w_t - \w^\star}| + 8 R^2\sum_{t=1}^T \|\bxi_t\|^2, \quad \text{(by the left-hand side inequality in \eqref{eq:lower})} \nn \\
& \leq 4 GR \sum_{t=1}^T \inner{\gb_t}{\w_t - \w^\star}+ 8 G^2 R^2 +  8 R^2\sum_{t=1}^T \|\bxi_t\|^2, \quad \text{(by \eqref{eq:lower2})}
\end{align}
Plugging this into \eqref{eq:pretogether2} and rearranging, we get 
\begin{align}
    \sum_{t=1}^T \inner{\g_t}{\w_t -\wtilde\w^\star} -  2 GR\eta \sum_{t=1}^T \inner{\gb_t}{\w_t - \w^\star} 
    & \leq 4 \eta G^2 R^2 +  4\eta R^2\sum_{t=1}^T \|\bxi_t\|^2 + 5 \kappa G R + \nu \log T+  \frac{d \log (1+\frac{T}{d})}{\eta}.
\end{align}
Taking the expectation on both sides and using that $\E[\g_t]=\gb_t$ and $\E[\|\bxi_t\|^2]\leq \sigma^2$, we get 
\begin{align}
    4 \eta R^2 T \sigma^2 +  4 \eta G^2 R^2  + 5 \kappa G R + \nu \log T + \frac{d \log (1+\frac{T}{d})}{\eta} & \geq   \E\left[\sum_{t=1}^T \inner{\g_t}{\w_t -\wtilde\w^\star}\right]-2GR\eta \cdot\E\left[\sum_{t=1}^T \inner{\g_t}{\w_t -\w^\star}\right] ,\nn \\
& \geq (1-2 GR \eta) \cdot \E\left[\sum_{t=1}^T \inner{\g_t}{\w_t -\w^\star}\right] - GR, \quad (\text{by \eqref{eq:wtildegregret}})\nn \\
& \geq \frac{T}{2} \cdot (\E[f(\what\w_T)] - f(\w^\star)), 
\end{align}
where the last inequality follows by the fact that $\eta \leq \frac{1}{4 GR}$ and \eqref{eq:thats}. Now, dividing by $\frac{T}{2}$ on both sides and rearranging, we get 
\begin{align}
    \E[f(\what\w_T)] - f(\w^\star)& \leq   8 \eta R^2  \sigma^2 + \frac{8 \eta G^2 R^2}{T}  + \frac{12 \kappa G R}{T} + \frac{2\nu \log T}{T} + \frac{2d \log (1+\frac{T}{d})}{\eta T},\nn \\
    & \leq  8 \eta R^2  \sigma^2   + \frac{2d \log (1+\frac{T}{d})}{\eta T}+ \frac{14 \kappa G R}{T} + \frac{2\nu \log T}{T},\label{eq:rep}
\end{align}
where the last inequality follows by $\eta\leq \frac{1}{10 \kappa GR}$. Note that the optimal tuning of $\eta$ in \eqref{eq:rep} is given by 
\begin{align}
    \eta^\star = \sqrt{\frac{d \log(1 + \frac{T}{d})}{4 R^2 \sigma^2 T }}.
\end{align}
We now consider cases. 
\fakepar{Case where $\eta^\star \leq \frac{1}{10 \kappa G R}$} First, note that this implies that $\eta = \eta^\star$. Now, using that $\eta^\star \leq \frac{1}{10 \kappa G R}$ and the expression of $\eta^\star$, we have that 
\begin{align}
    5 G \kappa \leq \sigma \sqrt{\frac{T}{d\log(1+ \frac{T}{d})}}.
\end{align}
This implies that $\nu \leq 4\sigma \sqrt{\frac{d T}{ \log (1+ \frac{T}{d})}}$ (see definition of $\nu$ in \eqref{eq:params2}). Using this together with $\eta =\eta^\star$ and \eqref{eq:rep} implies that 
\begin{align}
  (\text{case } \eta^\star \leq \tfrac{1}{10 \kappa GR}) \quad \E[f(\what\w_T)] -f(\w^\star)& \leq 8 R \sigma  \cdot \sqrt{\frac{d}{T}} + \frac{14 \kappa G R}{T} + 8 R \sigma \sqrt{\frac{d}{T  \log(1+\frac{T}{d})}} \cdot \log T,\nn \\
  & \leq 16 R \sigma  \cdot \sqrt{\frac{d \log (1+\frac{T}{d})}{T}} + \frac{  14 \kappa G R}{T}.\label{eq:bounding11}
\end{align}
\fakepar{Case where $\eta^\star \geq \frac{1}{10 \kappa G R}$} In this case, we have $\eta =\frac{1}{10 \kappa GR}$. Now, using that $\eta^\star\geq \frac{1}{10 \kappa GR}$ and the expression of $\eta^\star$, we have 
\begin{align}
    5 G\kappa\geq \sigma \sqrt{\frac{T}{d \log(1+ \frac{T}{d})}}.  \label{eq:vance}
\end{align}
This implies  that $\nu\leq 20 G R  \kappa d$ (see definition of $\nu$ in \eqref{eq:params2}). Plugging this and $\eta=\frac{1}{10 \kappa G R}$ into \eqref{eq:rep}, we get 
\begin{align}
    \text{(case $\eta^\star\geq \tfrac{1}{10 \kappa G R}$)}\quad	\E[f(\what\w_T)] -f(\w^\star)  &\leq \frac{4R  \sigma^2}{5\kappa G}   + \frac{20GR \kappa d \log (1+\frac{T}{d})}{ T}+ \frac{14 \kappa G R}{T} + \frac{40 G R \kappa d \log T}{T},\nn \\
    &\leq \frac{4R  \sigma^2}{5\kappa G}   + \frac{74 GR \kappa d \log (1+\frac{T}{d})}{T},\nn \\
    &\leq  4 R\sigma \sqrt{\frac{d \log (1+ \frac{T}{d})}{T}}  +\frac{74 G R\kappa d\log(1 +\frac{T}{d})}{T},\label{eq:bounding22}
\end{align}
where the last inequality follows by \eqref{eq:vance}. Thus, combining \eqref{eq:bounding11} and \eqref{eq:bounding22}, we get 
\begin{align}
    \E[f(\what\w_T)] -f(\w^\star) \leq 16 R \sigma \sqrt{\frac{d \log (1+ \tfrac{T}{d})}{T}}  +  \frac{74 G R\kappa d\log(1 +\frac{T}{d})}{T}.
\end{align}
This proves the desired convergence rate.

\paragraph{Computational cost} By \cref{thm:regretNewton} and the choice of $c=1/2$, the computational cost of the \barons{} subroutine with \cref{alg:projectionfreewrapper} is bounded by $\wtilde{O}\left( d^2 T +  d^{\omega} \sqrt{\tfrac{d T}{\nu \eta}} \right)$.
Now, by the choice of $\eta$ and $\nu$ in \eqref{eq:parameterchoice}, we have $\eta \nu\geq d$. This implies that the computation of the \barons{} subroutine is bounded by 
\[
\wtilde{O}\left(d^2 T +  d^{\omega}\sqrt{T}\right).
\]
Now, in addition to the computational cost of the \barons{} subroutine, \cref{alg:projectionfreewrapper} incurs $O(d) + C_{\gauged}(\cK)$ per round, where $C_{\gauged}(\cK)$ is the cost of one call to the $\gauged$ subroutine (\cref{alg:gauge}) for approximating the gauge distance $S_\K$ and its the subgradients. By \cref{lem:approxgrad}, we have 
\begin{align}
	C_{\gauged}(\cK) \leq \wtilde{O} (1)\cdot  C_{\sep}(\cK).
\end{align}
This implies the desired computational cost.
\end{proof}  
\section{Computing the Gauge Distance (Proof of \cref{lem:approxgrad})}
\label{sec:fullproof}
\label{sec:subgradient}
For the proof of \cref{lem:approxgrad}, we need the following properties of the Gauge function (see e.g.~\cite{molinaro2020curvature} for a proof).
\begin{lemma}
	\label{lem:properties2}
	Let $\w\in \reals^d\setminus \{\bm{0}\}$ and $0< r\leq R$. Further, let $\cC$ be a closed convex set such that $\cB(r) \subseteq \cC \subseteq \cB(R)$. Then, the following properties hold:
	\begin{enumerate}[label=\alph*.]
		\item $\gamma_{\cC}(\w) =\sigma_{\cC^\circ}(\w)= \sup_{\x\in \cC^\circ}\x^\top \w$ and $(\cC^\circ)^\circ =\cC$.
		\item $\sigma_{\cC}(\alpha \w) = \alpha \sigma_{\cC}(\w)$ and $\partial \sigma_{\cC}(\alpha \w)= \partial \sigma_{\cC}(\w) = \argmax_{\bu \in \cC} \inner{\bu}{\w}$, for all $\alpha\geq 0$.  %
		\item $r \|\w\|\leq \sigma_{\cC}(\w) \leq R\|\w\|$, $\|\w\|/R\leq \gamma_{\cC}(\w) \leq \|\w\|/r$, and $\cB(1/R)\subseteq \cC^\circ \subseteq \cB(1/r)$.
	\end{enumerate}
\end{lemma}
With this, we now prove \cref{lem:approxgrad}.
\begin{proof}[Proof of \cref{lem:approxgrad}]
  Fix $\w\in \reals^d$. We consider cases. If $\w\in \cC$, then the `if' condition in \cref{line:memset} of \cref{alg:gauge} evaluates to `true', and so the algorithm returns the pair $(S, \bs) = (0,\bm{0})$. Since $\w\in \cC$, we have $\gamma_\cC(\w) \leq 1$, and so by \cref{lem:gauge}, we have for all $\u\in \reals^d$: 
  \begin{align}
    S_\cC(\u)= \max(0,\gamma_\cC(\u)-1)\geq  0 = \max(0,\gamma_\cC(\w)-1)= S_\cC(\w).  
  \end{align}  
  This implies the desired result since $\bm{0}\in \partial S_\cC(\w)$ by \cref{lem:gauge}.

Now, consider the case where $\w\not\in \cC$, and let $\alpha$, $\beta$, $\mu$, $\vv$, and $\bs$ be as in \cref{alg:gauge} when the algorithm returns. Then, by design, when \cref{alg:gauge} returns, we have
\begin{align}
\alpha \w \in \cC, \quad \beta  \w \not\in \cC, \quad \text{and}\quad   |\beta - \alpha| \leq \frac{r^2\veps}{2 \|\w\|^2}. \label{eq:ff}
\end{align}
Since $\gamma_\cC(\w) = \inf\{\lambda >0\mid \w \in  \lambda \cC \}$, we have that 
\begin{align}
\frac{1}{\beta} \leq  \gamma_\cC(\w) \leq \frac{1}{\alpha}. \label{eq:athlet}
\end{align}
Now, since $\gamma_\cC(\w) \leq \|\w\|/r$ (by \cref{lem:properties2}.c), the left-hand side inequality in \eqref{eq:athlet} implies that
\begin{align}
\beta \geq \frac{r}{\|\w\|}.\label{eq:oneside}
\end{align}
Note that $\|\w\|>0$ since $\w\not\in\cC$ and $\bbB(r)\subseteq \cC$. Using \eqref{eq:oneside} and the fact that $|\beta - \alpha| \leq \frac{r^2\veps}{2\|\w\|^2}$ (see \eqref{eq:ff}), we have 
\begin{align}
\frac{1}{\alpha} & \leq \frac{1}{\beta - \frac{r^2\veps}{2\|\w\|^2}}, \nn \\
&\leq \frac{1}{\beta} + \frac{r^2 \veps}{\beta^2\|\w\|^2}, \quad (\text{see below}) \label{eq:ineqqq} \\
&\leq \frac{1}{\beta} + \veps, \quad \text{(by \eqref{eq:oneside})} \label{eq:cucial}
\end{align}
where \eqref{eq:ineqqq} follows by \eqref{eq:oneside} and the fact that $\frac{1}{1-x}\leq 1 + 2 x$, for all $x\leq \frac{1}{2}$; we instantiate the latter with $x= \frac{r^2 \veps}{2\beta\|\w\|^2}$ which satisfies $x\leq 1/2$ since $\|\w\|\geq r$ (because $\w\not\in \cC$ and $\bbB(r)\subseteq \cC$). Combining \eqref{eq:cucial} with \eqref{eq:athlet} and using that $S= \alpha^{-1}-1$ (see \cref{alg:gauge}), we get 
\begin{align}
  \gamma_\cC(\w)\leq S +1 \leq \gamma_\cC(\w) +\veps. \label{eq:leverage}
\end{align}
This together with the facts that $S_{\cC}(\w)=\max(0,\gamma_\cC(\w)-1)$ and $\gamma_\cC(\w)\geq 1$ (since $\w\not\in\cC$) implies that $S_\cC(\w)\leq S \leq S_\cC(\w) +\veps$, as desired.

\fakepar{Approximate subgradient} We now show that the second output $\bs$ of \cref{alg:gauge} is an approximate subgradient of the gauge distance function at $\w$.

Since $\vv$ is the separating hyperplane returned by the call to $\Sep_\cC(\mu \w)$, we have that
    \begin{align}
        \forall \u \in \cC,\quad  \u^\top \vv \leq  \mu \w^\top \vv.
    \end{align} 
    Thus, since the vector $\bs$ returned by \cref{alg:gauge} satisfies $\bs =\frac{\vv}{\beta \cdot\w^\top \vv }$ and $\mu =\frac{\alpha +\beta}{2}\leq \beta$, we have 
    \begin{align}
\forall \u \in \cC,\quad  \u^\top \bs \leq  \mu \w^\top \bs \leq 1. \label{eq:polar}
    \end{align}
    This implies that $\bs \in \cC^\circ$ (by definition of the polar set) and so by \cref{lem:properties2}.a, this implies that 
    \begin{align}
        \forall \u \in \reals^d,\quad  \bs^\top \u  \leq \sup_{\x\in \cC^\circ} \x^\top \u = \gamma_\cC(\u). \label{eq:firt}
    \end{align}
   On the other hand, combining \eqref{eq:cucial} with \eqref{eq:athlet}, we get 
   \begin{align}
    \gamma_\cC(\w) - \veps \leq \frac{1}{\beta} = \bs^\top \w,\label{eq:alert}
   \end{align}
   where the equality uses the expression of $\bs$. Combining \eqref{eq:firt} and \eqref{eq:alert} implies that 
   \begin{align}
    \forall \u \in \reals^d,\quad  \bs^\top(\u - \w) + \gamma_{\cC}(\w) - \veps \leq \gamma_\cC(\u).
   \end{align}
Thus, subtracting $1$ from both sides and using that $S_\cC(\w)=\gamma_\cC(\w) -1$ (since $\w\not\in \cC$), we get that
   \begin{align}
    \forall \u \in \reals^d,\quad  \bs^\top(\u - \w) + S_{\cC}(\w) - \veps &\leq \gamma_\cC(\u)-1,\nn \\
    & \leq \max(0, \gamma_{\cC}(\u)-1),\nn \\
    &= S_\cC(\u).
   \end{align}
   This shows the inequality on the right-hand side of \eqref{eq:pand}. Now, as mentioned earlier, \eqref{eq:polar} implies that $\bs\in \cC^\circ$. And since $\bbB(r) \subseteq \cC$, we have $\cC^\circ \subseteq \bbB(1/r)$ by \cref{lem:properties2}. Therefore, $\|\bs\|\leq 1/r$.

   \fakepar{Number of oracle calls}
   The number of oracle calls is bounded by the number of iterations of the `while' loop in \cref{line:while}. Since \cref{alg:gauge} implements a bisection, the number of iteration is at most $\log_2(\frac{4 \|\w\|^2}{r^2 \veps})$.
\end{proof}

\end{document}